# MPCFormer: A physics-informed data-driven approach for explainable socially-aware autonomous driving

Jia Hu, *Senior Member, IEEE*, Zhexi Lian, Xuerun Yan, Ruiang Bi, Dou Shen, Yu Ruan, Chunlong Xia, and Haoran Wang

***Abstract*— Autonomous Driving (AD) vehicles still struggle to exhibit human-like behavior in highly dynamic and interactive traffic scenarios. The key challenge lies in AD's limited ability to interact with surrounding vehicles, largely due to a lack of understanding the underlying mechanisms of social interaction. To address this issue, we introduce *MPCFormer*, an explainable socially-aware autonomous driving approach with physics-informed and data-driven coupled social interaction dynamics. In this model, the dynamics are formulated into a discrete space-state representation, which embeds physics priors to enhance modeling explainability. The dynamics coefficients are learned from naturalistic driving data via a Transformer-based encoder-decoder architecture. To the best of our knowledge, MPCFormer is the first approach to explicitly model the dynamics of multi-vehicle social interactions. The learned social interaction dynamics enable the planner to generate manifold, human-like behaviors when interacting with surrounding traffic. By leveraging the MPC framework, the approach mitigates the potential safety risks typically associated with purely learning-based methods. Open-looped evaluation on NGSIM dataset demonstrates that MPCFormer achieves superior social interaction awareness, yielding the lowest trajectory prediction errors compared with other state-of-the-art approaches. The prediction achieves an ADE as low as 0.86 m over a long prediction horizon of 5 seconds. Close-looped experiments in highly intense interaction scenarios, where consecutive lane changes are required to exit an off-ramp, further validate the effectiveness of MPCFormer. Results show that MPCFormer achieves the highest planning success rate of 94.67%, improves driving efficiency by 15.75%, and reduces the collision rate from 21.25% to 0.5%, outperforming a frontier Reinforcement Learning (RL) based planner.***

This paper is partially supported by National Natural Science Foundation of China (Grant No. 52372317 and 52302412), Yangtze River Delta Science and Technology Innovation Joint Force (No. 2023CSJGG0800), Shanghai Automotive Industry Science and Technology Development Foundation (No. 2404), Xiaomi Young Talents Program, the Fundamental Research Funds for the Central Universities (22120230311), Tongji Zhongte Chair Professor Foundation (No. 000000375-2018082), and Shanghai Sailing Program (No. 23YF1449600).(*Corresponding author: Haoran Wang*)

Jia Hu, Zhexi Lian, Xuerun Yan, Ruiang Bi, and Haoran Wang are with Key Laboratory of Road and Traffic Engineering of the Ministry of Education, College of Transportation, Tongji University, No.4800 Cao'an Road, Shanghai, PR China, 201804. (e-mail: hujia@tongji.edu.cn,zhexi_lian@tongji.edu.cn,frankyan@tongji.edu.cn, 2431720@tongji.edu.cn, wang_haoran@tongji.edu.cn).

Dou Shen, Yu Ruan, and Chunlong Xia are with Beijing Baidu Netcom Science Technology Co., Ltd, No. 8 Northeast Wangxi Road, Beijing, PR China, 100006. (e-mail: shendou@baidu.com, ruanyu@baidu.com, xiachunlong@baidu.com).

## I. Introduction

### A. *Research motivation*

During recent years, Autonomous Driving (AD) has demonstrated significant progress within transportation systems [1][2]. However, AD vehicles still face significant challenges in exhibiting human-like behavior in highly dynamic and interactive traffic scenarios such as off-ramp and unprotected left turns [3][4]. One critical reason is that AD vehicles lack the understanding of the underlying mechanisms of social interaction between surrounding vehicles. To be detailed, AD vehicles are not sure how surrounding vehicles will behave and how to react to the effects of surrounding vehicles' behaviors. Given an illustrative example in Figure 1 (a), the ego AD vehicle attempts a lane change. But a lack of social awareness makes it fail to understand and adapt to the motion cues of the surrounding vehicle [5], resulting in high safety risks. By contrast, in Figure 1 (b), a skilled human driver can recognize the potential behavior consequence of the surrounding vehicle and give up the maneuver for safety. Hence, there is a great need to develop AD vehicles with socially-aware driving capability.

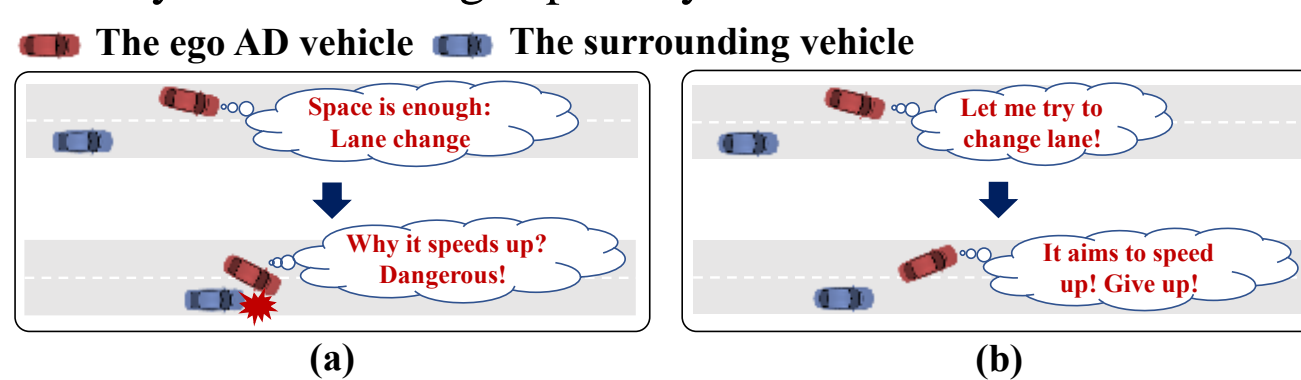

Figure 1 A lane-change maneuver comparison example between (a) the AD vehicle and (b) a skilled human driver.

### B. *Related work and limitations*

Current mainstream socially-aware autonomous driving approaches can be divided into three categories based on the development process: "passively", "neutrally", and "proactively" socially-aware approaches.

To enable the AD vehicle to interact effectively with other vehicles, early researchers developed PAssively Socially-aware (PAS) approaches as shown in Figure 2 (a). This type of approach usually follows a hierarchical "prediction → reaction" fashion, which means predicting other vehicles' trajectories first and passively reacting to the predicted trajectories secondly [6][7][8]. Consequently, to ensure safety in practice, the planned motions must avoid any conflicts with the predicted trajectories of other vehicles [9][10][11]. For instance, Wei et al [12] proposed a bi-level optimization framework grounded in game theory, wherein the AD vehicle's motion planning is conditioned on the predictions of

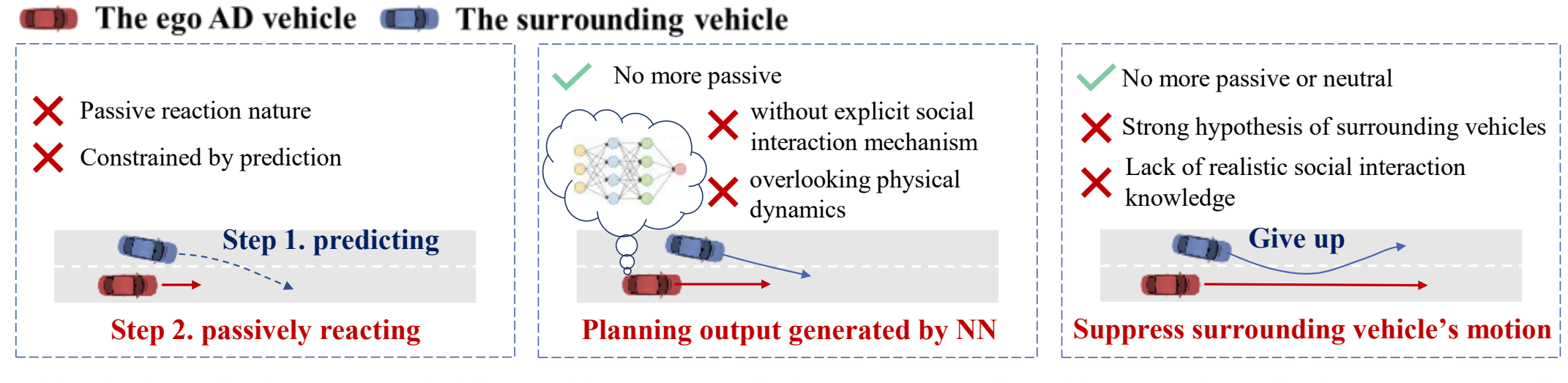

**(a) PAssively Socially-aware (PAS)** **(b) Neutrally Socially-aware (NS)** **(c) PRoactively Socially-aware (PRS)**

Figure 2 The illustration of current mainstream approaches and limitations.

TABLE 1 A detailed methodology comparison of existing socially-aware driving approaches.

| Type | Research | Category | Applied model | Explicit SI | Physics-informed | Data-driven |
|---|---|---|---|---|---|---|
| PAS | Sheng et al [9] | Search | POMDP | × | × | × |
| | Hubmann et al [10] | Search | POMDP+MCTS | × | × | × |
| | Zhou et al [8] | Optimization | MPC | × | √ | × |
| | Huang et al [6] | Learning | Transformer + QP | × | √ | √ |
| | Liao et al [11] | Learning | PINN Transformer | × | √ | √ |
| NS | Bronstein et al [16] | Learning | Transformer+GMM | × | × | √ |
| | Ye et al [17] | Learning | DRL | × | × | √ |
| | Hu et al [18] | Learning | Transformer + DRL | × | × | √ |
| | Liang et al [21] | Learning | Transformer + TL | × | × | √ |
| | Wang et al [22] | Learning | Transformer + RF | × | × | √ |
| | Tan et al [23] | Learning | Transformer + FM | × | × | √ |
| PRS | Gharavi et al [25] | Optimization | SMPC | × | √ | × |
| | Hu et al [26] | Game theory | IRL+MPC | √ | √ | × |
| | Liu et al [28] | Learning | Transformer + BT | √ | × | √ |
| | Huang et al [29] | Learning | Transformer + TS | × | √ | √ |
| | Yan et al [30] | Learning | MM Transformer | √ | × | √ |
| | **Ours MPCFormer** | Learning | PI Transformer + GMPC | √ | √ | √ |

Notes: POMDP--Partially Observed Markov Decision Process; MCTS--Monte Carlo Tree Search; MPC--Model Predictive Control; QP--Quadratic Programming; PINN--Physics-informed Neural Network; GMM--Gaussian Mixture Model; DRL--Deep Reinforcement Learning; TL--Transfer Learning; RF--Risk Field; FM--Flow Matching; SMPC--Stochastic MPC; IRL--Inverse RL; HRL--Human-guided RL; BT--Behavior Topology; TS--Tree Structure; MM--Multimodal; PI--Physics Informed; GMPC--Game MPC.

surrounding vehicles. In other words, the AD vehicles are constrained by predictions and have to yield to surrounding vehicles. Hence, the primary limitation of PAS approaches lies in their passive reaction nature, failing to proactively account for the behaviors of other vehicles. The generated motion planning trajectory must adhere to constraints imposed by the predicted trajectories of other vehicles. These constraints significantly narrow the solution space for motion planning, ultimately resulting in reduced planning effectiveness.

To overcome the passivity of PAS approaches, researchers have proposed Neutrally Socially-aware (NS) approaches as shown in Figure 2 (b). AD vehicles no longer react passively to surrounding vehicles in these approaches. Instead, they adopt an "environmental states input→Neural Networks (NN) → planning output" framework [13][14]. Through deep learning techniques, AD vehicles learn how to interact with surrounding vehicles and select motions based solely on the current and historical environmental states. A key contribution is made by Chitta et al [15]. They used perception-fused attention blocks and an autoregressive decoder for planned motion generation. However, two limitations still exist in NI approaches. Firstly, these approaches cannot explicitly explain the social interaction mechanism between AD vehicles and surrounding vehicles. These approaches rely on black-box models [16][17], which just fit mapping laws from input to output [22][23] but fail to provide clear insights into how vehicles interact with each other [18][21][24]. Secondly, these approaches often overlook the underlying physical principles that govern vehicle dynamics. As these approaches often lack inherent consideration of vehicle dynamics [19][20], they potentially generate sudden and sharp maneuvers which exceed the vehicle's operational capabilities.

More recently, fewer researchers proposed PRoactively Socially-aware (PRS) approaches, as shown in Figure 2 (c). In order to achieve driving goals, these approaches can suppress the surrounding vehicles' behaviors through proactive motions [25][29][30]. For instance, the ego AD vehicle could accelerate to narrow the following gap when surrounding vehicles attempt to cut in [26]. This proactive maneuver could discourage the potential cut-ins. While effective, one limitation in common has been raised. These approaches usually give strong behavioral hypotheses of surrounding vehicles' behavior [28] and lack inherent knowledge of the social interaction mechanism. For instance, Hu et al [26] developed a proactive motion planner with surrounding vehicles' behavior consideration, and they assumed that surrounding vehicles are governed by a multi-objective cost function. However, when there is a significant behavior difference between the surrounding vehicles and the hypothesis, these approaches may fail to account for changes

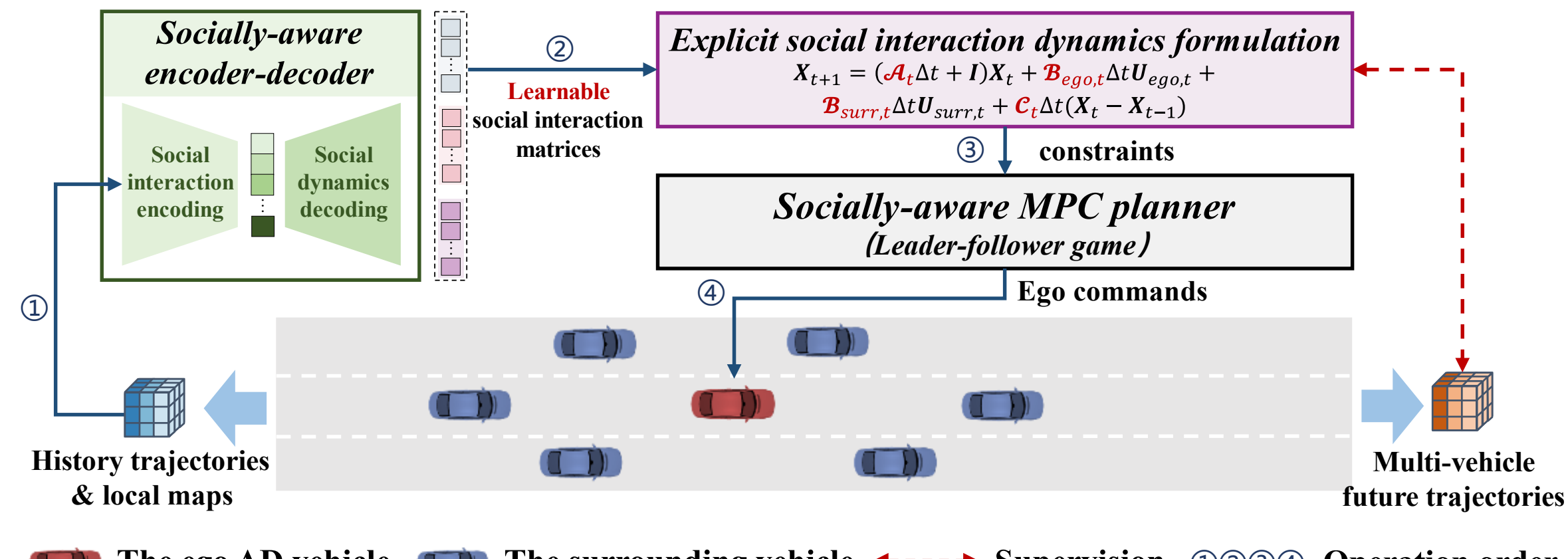

Figure 3 The overall architecture of the proposed approach. The traffic consists of ego AD vehicle (EAV) and six surrounding vehicles (SVs), i.e., front vehicle (FV), rear vehicle (RV), left-front vehicle (LFV), left-rear vehicle (LRV), right-front vehicle (RFV), and right-rear vehicle (RRV)

in the behavior of certain surrounding vehicles. This may lead to unforeseen conflicts or an inability to effectively avoid potential hazards [27][31].

### C. Contributions

To this end, we introduce ***MPCFormer***, an explainable socially-aware autonomous driving approach with physics-informed and data-driven coupled social interaction dynamics. In this model, the dynamics are formulated into a discrete space-state representation, which embeds physics priors to enhance modeling explainability. The dynamics coefficients are learned from naturalistic driving data. Different from existing learning-based approaches (i.e. Transformer-based), MPCFormer is the first approach to explicitly learn physics-informed model dynamics of multi-vehicle social interactions. By integrating the learned social interaction dynamics into MPC planning based on a leader-follower game framework, the AD vehicle can generate manifold, human-like behaviors when interacting with surrounding traffic while mitigating the potential safety risks typically associated with purely learning-based approaches. The main contributions of the proposed approach are listed as follows.

***a) Our approach is the first to explicitly model the dynamics of multi-vehicle social interactions through a physics-informed data-driven modeling paradigm.*** The proposed approach explicitly models the dynamics of social interactions among vehicles, termed the explicit social interaction dynamics. To ensure explainability, we propose a physics-informed and data-driven coupled modeling paradigm. This paradigm formulates a state-space representation formula coupling behaviors of both the ego AD vehicle and the surrounding vehicles. Among the socially-aware system dynamics, each vehicle's state transition process is influenced not only by its own kinematic dynamics but also by social interaction effects from surrounding vehicles' behaviors. Hence, two types of components are integrated in the modeling paradigm: (1) physics-informed components, which reflect vehicles' inherent physical kinematic dynamics; (2) data-driven components, which reflect the social interaction effects between multiple vehicles. The data-driven components are explicitly modeled as learnable matrices, enabling them to capture and explain complex inter-vehicle social interaction mechanism. To learn the data-driven components, we design a socially-aware Transformer-based encoder-decoder that fully exploits the social interaction mechanism embedded in natural driving data.

***b) Our approach enhances proactively socially-aware autonomous driving with explicit social interaction dynamics consideration.*** The proposed approach integrates the explicit social interaction dynamics into the ego AD vehicle's MPC planning under a leader-follower game framework, enabling it to anticipate how various actions will trigger different reactions from surrounding vehicles. Hence, the ego AD vehicle can proactively plan its motions to achieve driving goals, while considering surrounding vehicles' all possible reactions.

***c) Our approach enables joint trajectory prediction and motion planning.*** The proposed approach can integrate prediction into motion planning by setting the explicit social interaction dynamics as constraints in the MPC-based motion planning process. Hence, when optimizing motions of the ego AD vehicle within the future planning horizon, the future states of surrounding vehicles are also predicted simultaneously through the explicit social interaction dynamics. Hence, trajectory prediction and motion planning can be accomplished jointly.

## II. Methodology

### A. Overall architecture

The overall architecture is shown in Figure 3. It consists of three modules:

- **Socially-aware encoder-decoder:** this module utilizes trajectories and local maps of the ego AD vehicle and surrounding vehicles to generate the learnable dynamics coefficients (also called social interaction matrices) formulated in the explicit social interaction dynamics (Section II. B). The inter-vehicle social interactions are fully accounted for in the Transformer-based encoder-decoder structure. The structure of the encoder-decoder is detailed in Section II. C.

- **Explicit social interaction dynamics formulation:** this module takes the learned social interaction matrices and formulates the aforementioned explicit social interaction dynamics, in which both physics-informed vehicle kinematic and learnable inter-vehicle social interactions are coupled modeled. The explicit social interaction dynamics serve as constraints in the ego AD vehicle motion planning. The formulation process is detailed in Section II. B.

- **Socially-aware MPC planner:** this module designs an MPC-based motion planner for the ego AD vehicle. The planner can generate socially-aware behaviors through taking the explicit social interaction dynamics as the social interaction constraint based on a leader-follower game framework. It also mitigates the potential safety risks through hard collision avoidance constraints. The design of the planner is detailed in Section II. D.

### B. *Explicit social interaction dynamics formulation*

This section formulates the explicit social interaction dynamics. The social interaction effects between vehicles are considered as data-driven components in the dynamics and explicitly modeled as learnable matrices.

#### 1) *Preliminary: single vehicle dynamics modeling*

The single vehicle dynamics serve as the state transition model. Since EAV and SVs conduct longitudinal and lateral coupled maneuvers, the state vector is defined as:

$$\boldsymbol{x}_{ego} = \left(s_{ego}, v_{ego}, y_{ego}, \psi_{ego}\right)^{\mathrm{T}} \tag{1}$$

$$\boldsymbol{x}_{surr_i} = \left(s_{surr_i}, v_{surr_i}, y_{surr_i}, \psi_{surr_i}\right)^{\mathrm{T}} \tag{2}$$

where $\boldsymbol{x}_{ego}$ denotes the state of EAV and $\boldsymbol{x}_{surr_i}$ denotes the state of SV $i$; $s_{ego}$ and $s_{surr_i}$ are longitudinal positions of EAV and SV $i$ under the Frenet coordinate; $v_{ego}$ and $v_{surr_i}$ represent the speed of EAV and SV $i$; $y_{ego}$ and $y_{surr_i}$ are lateral positions of EAV and SV $i$ under the Frenet coordinate; $\psi_{ego}$ and $\psi_{surr_i}$ are the heading angles of EAV and SV $i$. The control inputs of EAV and SVs are defined as follows. It is worth noting that only EAV can be controlled proactively.

$$\boldsymbol{u}_{ego} = \left(a_{ego}, {\delta_f}_{ego}\right)^{\mathrm{T}} \tag{3}$$

$$\boldsymbol{u}_{surr_i} = \left(a_{surr_i}, {\delta_f}_{surr_i}\right)^{\mathrm{T}} \tag{4}$$

In Eq. (3) and (4), $\boldsymbol{u}_{ego}$ denotes the control input of EAV and $\boldsymbol{u}_{surr_i}$ denotes the control input of SV $i$; $a_{ego}$ and $a_{surr_i}$ are the accelerations of EAV and SV $i$; ${\delta_f}_{ego}$ and ${\delta_f}_{surr_i}$ are the front steering angles of EAV and SV $i$.

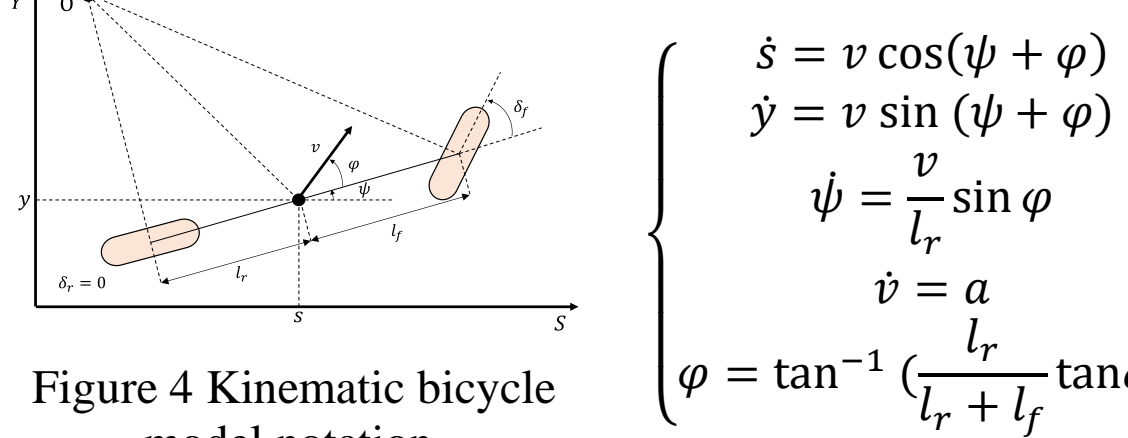

Figure 4 Kinematic bicycle model notation.

$$\begin{cases} \dot{s} = v\cos(\psi + \varphi) \\ \dot{y} = v\sin(\psi + \varphi) \\ \dot{\psi} = \dfrac{v}{l_r}\sin\varphi \\ \dot{v} = a \\ \varphi = \tan^{-1}\left(\dfrac{l_r}{l_r + l_f}\tan\delta_f\right) \end{cases} \tag{5}$$

The vehicle system dynamics are modeled using the vehicle kinematic bicycle model [32]. Compared with lateral-only or longitudinal-only models, the adopted kinematic bicycle model captures the coupled longitudinal–lateral vehicle motion, which can involve both speed adjustment and lane-changing behaviors. Meanwhile, it maintains a good balance between modeling fidelity and computational efficiency. As illustrated in Figure 4 and Eq. (5), $\varphi$ is the angle of the current speed of gravity center with respect to the longitudinal axis of the vehicle; $l_r$ and $l_f$ are distance between the vehicle's gravity center and the rear and front axles, respectively. Using the small angle assumption of $\varphi$, the kinematic bicycle model in Eq. (5) is linearized. Subsequently, the linear vehicle system dynamics, which incorporates the kinematic models, can be derived as follows:

$$\dot{\boldsymbol{x}}_{ego} = \boldsymbol{A}_{ego}\boldsymbol{x}_{ego} + \boldsymbol{B}_{ego}\boldsymbol{u}_{ego} \tag{6}$$

$$\dot{\boldsymbol{x}}_{surr_i} = \boldsymbol{A}_{surr_i}\boldsymbol{x}_{surr_i} + \boldsymbol{B}_{surr_i}\boldsymbol{u}_{surr_i} \tag{7}$$

where,

$$\boldsymbol{A}_{ego} = \begin{bmatrix} 0 & 1 & 0 & 0 \\ 0 & 0 & 0 & 0 \\ 0 & 0 & 0 & v_{ego} \\ 0 & 0 & 0 & 0 \end{bmatrix}, \boldsymbol{A}_{surr_i} = \begin{bmatrix} 0 & 1 & 0 & 0 \\ 0 & 0 & 0 & 0 \\ 0 & 0 & 0 & v_{surr_i} \\ 0 & 0 & 0 & 0 \end{bmatrix} \tag{8}$$

$$\boldsymbol{B}_{ego} = \begin{bmatrix} 0 & 0 \\ 1 & 0 \\ 0 & 0 \\ 0 & \dfrac{v_{ego}}{l_r + l_f} \end{bmatrix}, \boldsymbol{B}_{surr_i} = \begin{bmatrix} 0 & 0 \\ 1 & 0 \\ 0 & 0 \\ 0 & \dfrac{v_{surr_i}}{l_r + l_f} \end{bmatrix} \tag{9}$$

#### 2) *Explicit social interaction dynamics modeling*

The explicit social interaction dynamics incorporate **social interaction effects** between EAV and SVs. For EAV and its SVs, the generalized state vector is defined as:

$$\boldsymbol{X} = \left(\boldsymbol{x}_{ego}, \boldsymbol{x}_{surr_1}, \dots, \boldsymbol{x}_{surr_n}\right)^{\mathrm{T}} \tag{10}$$

where $\boldsymbol{X}$ is the system's state vector, which is a set of single vehicles' state vectors. The control input of the explicit social interaction dynamics can also be generalized from Eq. (2):

$$\boldsymbol{U}_{ego} = \Big(\boldsymbol{u}_{ego}, \underbrace{\boldsymbol{0}, \dots, \boldsymbol{0}}_{n}\Big)^{\mathrm{T}} \tag{11}$$

$$\boldsymbol{U}_{surr} = \left(0, \boldsymbol{u}_{surr_1}, \dots, \boldsymbol{u}_{surr_n}\right)^{\mathrm{T}} \tag{12}$$

where $\boldsymbol{U}_{ego}$ and $\boldsymbol{U}_{surr}$ denotes control input of EAV and SVs respectively. $\boldsymbol{U}_{surr}$ can be seen as reactions because it can be considered that SVs react to the EAV's motions. Then, the explicit social interaction dynamics incorporate data-driven social interaction components between EAV and SVs. Our discrete-time modeling paradigm is as follows:

$$\boldsymbol{X}_{t+1} = (\boldsymbol{\mathcal{A}}_t\Delta t + \boldsymbol{I})\boldsymbol{X}_t + \boldsymbol{\mathcal{B}}_{ego,t}\Delta t\boldsymbol{U}_{ego,t} + \boldsymbol{\mathcal{B}}_{surr,t}\Delta t\boldsymbol{U}_{surr,t} + \boldsymbol{\mathcal{C}}_t\Delta t(\boldsymbol{X}_t - \boldsymbol{X}_{t-1}) \tag{13}$$

where,

$$\boldsymbol{\mathcal{A}}_t = \begin{bmatrix} \boldsymbol{A}_{ego,t} & \boldsymbol{0} & \cdots & \boldsymbol{0} \\ \boldsymbol{0} & \boldsymbol{A}_{surr_1,t} & \cdots & \boldsymbol{0} \\ \vdots & \vdots & \ddots & \vdots \\ \boldsymbol{0} & \boldsymbol{0} & \cdots & \boldsymbol{A}_{surr_n,t} \end{bmatrix} \tag{14}$$

$$\boldsymbol{\mathcal{B}}_{ego,t} = \begin{bmatrix} \boldsymbol{B}_{ego,t} & \boldsymbol{0} & \cdots & \boldsymbol{0} \\ \boldsymbol{B}_{surr_1,ego,t} & \boldsymbol{0} & \cdots & \boldsymbol{0} \\ \vdots & \vdots & \ddots & \vdots \\ \boldsymbol{B}_{surr_n,ego,t} & \boldsymbol{0} & \cdots & \boldsymbol{0} \end{bmatrix} \tag{15}$$

$$\boldsymbol{\mathcal{B}}_{surr,t} = \begin{bmatrix} \boldsymbol{0} & \boldsymbol{B}_{ego,surr_1,t} & \cdots & \boldsymbol{B}_{ego,surr_n,t} \\ \boldsymbol{0} & \boldsymbol{B}_{surr_1,t} & \cdots & \boldsymbol{B}_{surr_1,surr_n,t} \\ \vdots & \vdots & \ddots & \vdots \\ \boldsymbol{0} & \boldsymbol{B}_{surr_n,surr_1,t} & \cdots & \boldsymbol{B}_{surr_n,t} \end{bmatrix} \tag{16}$$

$$\boldsymbol{\mathcal{C}}_t = \begin{bmatrix} \mathbf{0} & \boldsymbol{C}_{ego,surr_1,t} & \cdots & \boldsymbol{C}_{ego,surr_n,t} \\ \boldsymbol{C}_{surr_1,ego,t} & \mathbf{0} & \cdots & \boldsymbol{C}_{surr_1,surr_n,t} \\ \vdots & \vdots & \ddots & \vdots \\ \boldsymbol{C}_{surr_n,ego,t} & \boldsymbol{C}_{surr_n,surr_1,t} & \cdots & \mathbf{0} \end{bmatrix} \quad (17)$$

In Eq. (13), $\Delta t$ denotes the discrete time interval; $\boldsymbol{I}$ denotes a diagonal unit matrix. The subscript $t$ of each variable denotes the variable at time step $t$. It can be seen that the diagonal elements $\boldsymbol{A}_{ego,t}, \boldsymbol{A}_{surr_1,t}, \dots, \boldsymbol{A}_{surr_n,t}$ of the state matrix $\boldsymbol{\mathcal{A}}_t$ are defined by the kinematic bicycle model, as illustrated in Eq. (8). These diagonal matrices reflect the vehicle's intrinsic kinematic properties. The same applies to diagonal matrices of $\boldsymbol{\mathcal{B}}_{ego,t}$ and $\boldsymbol{\mathcal{B}}_{surr,t}$ in Eq. (15) and Eq. (16). These matrices are the physics-informed vehicle kinematic component.

In Eq. (15-17), the off-diagonal matrices explicitly reflect social interaction effects between vehicles. For instance, $\boldsymbol{C}_{ego,surr_1,t}$ denotes the social interaction effect of SV $i$ on EAV. The same applies to the control matrix $\boldsymbol{\mathcal{B}}_{ego,t}$ and $\boldsymbol{\mathcal{B}}_{surr,t}$ as well. These matrices are learned from natural driving data. Hence, the elements in $\boldsymbol{\mathcal{A}}_t, \boldsymbol{\mathcal{B}}_{ego,t}, \boldsymbol{\mathcal{B}}_{surr,t}$, and $\boldsymbol{\mathcal{C}}_t$ can be classified into two categories which are physics-informed kinematic matrices and data-driven social interaction matrices, as shown in TABLE 2. The physics-informed kinematic matrices are pre-defined and the data-driven social interaction matrices need to be learned. Though the dynamics are written in a linear state-space form, it includes interaction effects of each surrounding vehicle, and the social interaction matrices are generated by a nonlinear Transformer-based architecture. Thus, complex interactions are learned nonlinearly and then projected into a structured form.

TABLE 2 State and control matrices classification.

| | |
|---|---|
| Physics-informed kinematic matrices | $\boldsymbol{A}_{ego}, \boldsymbol{B}_{ego}$<br>$\boldsymbol{A}_{surr_1}, \boldsymbol{B}_{surr_1}$<br>…<br>$\boldsymbol{A}_{surr_n}, \boldsymbol{B}_{surr_n}$ |
| Data-driven social interaction matrices | $\boldsymbol{C}_{ego,surr_1,t}, \dots, \boldsymbol{C}_{ego,surr_n,t}, \boldsymbol{B}_{ego,surr_1,t}, \dots, \boldsymbol{B}_{ego,surr_n,t}$<br>$\boldsymbol{C}_{surr_1,ego,t}, \dots, \boldsymbol{C}_{surr_1,surr_n,t}, \boldsymbol{B}_{surr_1,ego,t}, \dots, \boldsymbol{B}_{surr_1,surr_n,t}$<br>…<br>$\boldsymbol{C}_{surr_n,ego,t}, \dots, \boldsymbol{C}_{surr_n,surr_{n-1},t}, \boldsymbol{B}_{surr_n,ego,t}, \dots, \boldsymbol{B}_{surr_n,surr_r}$ |

### C. Socially-aware encoder-decoder

In this section, this paper introduces how to learn the data-driven social interaction matrices in the explicit social interaction dynamics. We design a socially-aware Transformer-based encoder-decoder that fully exploits the social interaction mechanism between vehicles embedded in natural driving data. The whole structure of the socially-aware encoder-decoder is depicted in Figure 5.

#### 1) Input representation

The vehicle trajectory inputs contain the history trajectories of the EAV and SVs, defined as $\boldsymbol{Traj}_{veh,t} \in \mathbb{R}^{T_h \times f_{veh}}, veh \in \{EAV, FV, RV, LFV, LRV, RFV, RRV\}$ , where $T_h$ and $f_{veh}$ denote the number of historical timesteps before current time step $t$ and the number of trajectory features. The trajectory features include vehicle location, speed, acceleration, heading, and so on. The map inputs contain the map information in vehicle coordinates, termed $\boldsymbol{Map}_{veh,t} \in \mathbb{R}^{3 \times L \times f_{map}}, veh \in \{EAV, FV, RV, LFV, LRV, RFV, RRV\}$, where $L$ and $f_{map}$ denote the number of nearby waypoints and map features. The number 3 in the first dimension means three lanes are considered, which are the current lane, left lane, and right lane of the vehicle.

#### 2) Social Interaction encoding

**Vehicle-to-vehicle (V2V) social interaction encoding:** To encode the vehicle-to-vehicle social interaction, this paper firstly captures temporal attentions of the single vehicle's trajectory through TrajectoryFormer in Figure 5. The TrajectoryFormer mainly relies on an ensemble of self-attention blocks:

$$\widehat{\boldsymbol{Traj}}_{veh,t} = \mathrm{SelfAttn}\begin{pmatrix} \mathrm{MLP}(\boldsymbol{Traj}_{veh,t}), \mathrm{MLP}(\boldsymbol{Traj}_{veh,t}), \\ \mathrm{MLP}(\boldsymbol{Traj}_{veh,t}) \end{pmatrix}$$
$$veh \in \{EAV, FV, RV, LFV, LRV, RFV, RRV\} \quad (18)$$

$\widehat{\boldsymbol{Traj}}_{veh,t}$ is the processed temporal attention at time step $t$. SelfAttn is an ensemble of self-attention blocks, and MLP is a sequence of multi-layer perceptrons. The query, key, and value in the self-attention are all trajectory embeddings processed by MLP which aims to capture interaction effects among the trajectory set. Then this paper utilizes V2V Social Interaction Encoder to capture social interaction between

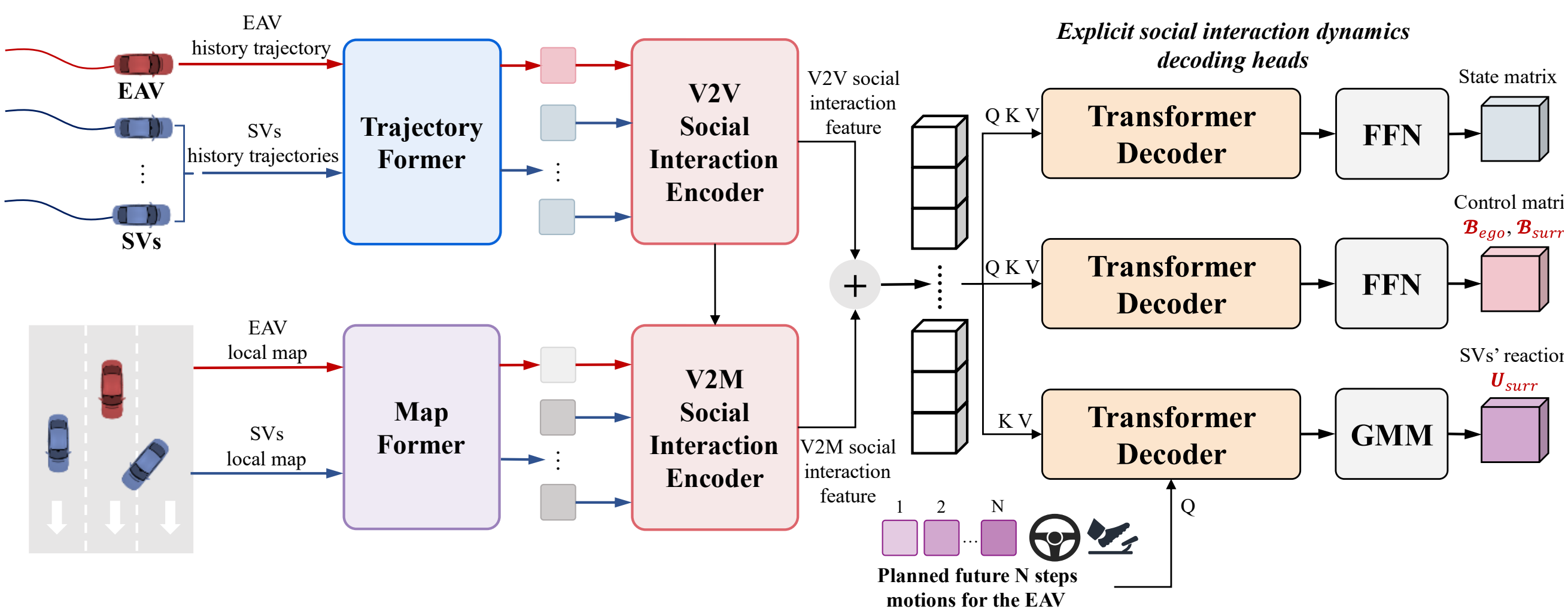

Figure 5 Illustration of the proposed socially-aware encoder-decoder

temporal attentions of vehicles. This encoder is composed of an ensemble of cross-attention blocks CrossAttn:

$$\boldsymbol{V2V}_{veh_1\to veh_2,t} = CrossAttn\begin{pmatrix}\widetilde{\boldsymbol{Traj}}_{veh_1,t}, \widetilde{\boldsymbol{Traj}}_{veh_2,t}, \\ \widetilde{\boldsymbol{Traj}}_{veh_2,t}\end{pmatrix} \tag{19}$$

$$veh_1, veh_2 \in \{EAV, FV, RV, LFV, LRV, RFV, RRV\}, veh_1 \neq veh_2$$

$\boldsymbol{V2V}_{veh_1\to veh_2,t}$ denotes the social interaction feature reflecting interaction effect of $veh_1$ on $veh_2$ at time step $t$. $veh_1$ and $veh_2$ denote any two different vehicles in EAV and SVs. The encoding is performed pairwise among all vehicles. It needs to be noted that this paper uses queries from $veh_1$ to capture the social interaction effect on $veh_2$ through choosing keys and values from $veh_2$.

**Vehicle-to-map (V2M) social interaction encoding:** To encode the vehicle-to-map social interaction, this paper firstly processes the map information vector $\boldsymbol{Map}_{veh}$ to embeddings using MLP, which forms MapFormer in Figure 5. Then this paper obtains the V2M social interaction feature through V2M Social Interaction Encoder. This encoder is also composed of an ensemble of cross-attention blocks:

$$\boldsymbol{V2M}_{veh_1\to veh_2,t} = \mathrm{CrossAttn}\begin{pmatrix}\boldsymbol{V2V}_{veh_1\to veh_2,t}, \\ \mathrm{MLP}\left(\boldsymbol{Map}_{veh_1,t}\right), \\ \mathrm{MLP}\left(\boldsymbol{Map}_{veh_1,t}\right)\end{pmatrix} \tag{20}$$

$$veh_1, veh_2 \in \{EAV, FV, RV, LFV, LRV, RFV, RRV\}, veh_1 \neq veh_2$$

$\boldsymbol{V2M}_{veh_1\to veh_2,t}$ denotes the V2M social interaction feature reflecting interaction effect of $veh_1$ on the local map of $veh_2$ at time step $t$. To capture the V2M social interaction, this paper utilizes the V2V social interaction feature $\boldsymbol{V2V}_{veh_1\to veh_2,t}$ as the queries and the map embeddings as the keys and values. That's because using the V2V social interaction feature as the query can fully investigate the relationship between map keys and values.

**Latent representation:** The latent representation $\boldsymbol{\mathcal{Z}}_t$ is obtained after the encoding process. It can be defined as:

$$\boldsymbol{\mathcal{Z}}_t = \mathrm{concat}\left(\left\{\mathbb{F}_{veh_1\to veh_2,t}\right\}\right) \tag{21}$$

$$veh_1, veh_2 \in \{EAV, FV, RV, LFV, LRV, RFV, RRV\}, veh_1 \neq veh_2$$

$$\mathbb{F}_{veh_1\to veh_2,t} = \mathrm{concat}\left(\boldsymbol{V2V}_{veh_1\to veh_2,t}, \boldsymbol{V2M}_{veh_1\to veh_2,t}\right) \tag{22}$$

$$veh_1, veh_2 \in \{EAV, FV, RV, LFV, LRV, RFV, RRV\}, veh_1 \neq veh_2$$

where $\mathbb{F}_{veh_1\to veh_2,t}$ denotes the aggregated social interaction feature reflecting social interaction effect of $veh_1$ on $veh_2$.

*3) Explicit social interaction dynamics decoding*

This paper designs three decoding heads as shown in Figure 5. The first and the second decoders are used to compute the data-driven social interaction matrices in state matrix $\boldsymbol{\mathcal{C}}$, control matrix $\boldsymbol{\mathcal{B}}_{ego}$ and $\boldsymbol{\mathcal{B}}_{surr}$ respectively (modeled in Section II.B) in the future planning horizon $N$. They are designed using typical Transformer Decoder Layers [33] as illustrated in Eq. (23) and (24).

The third decoder is used to compute the SVs' reactions $\boldsymbol{U}_{surr}$. Knowing that SVs react to the motions of EAV, this paper takes the planned future $N$ steps motions of the EAV as the queries in the Transformer Decoder Layer to query the latent representation for SVs' reaction predictions. After that, this paper uses Gaussian Mixture Model (GMM) to compute $M$ modalities of $\boldsymbol{U}_{surr}$, with average $\hat{\mu}_{\boldsymbol{U}_{surr}}$, variance $\hat{\sigma}_{\boldsymbol{U}_{surr}}$ and probability $\hat{p}$ of each modality. This paper selects the modality with the highest probability $\hat{p}$ rather than a weighted average of likely reactions, because it could potentially reduce the model's responsiveness to dominant interactions and cannot capture the most likely and impactful behavior.

In summary, at each time step $t$, the operation of the socially-aware encoder-decoder can be abstracted as—Given all vehicles' history trajectories, all vehicles' local maps, and EAV's planned motions to compute data-driven social interaction matrices of $\boldsymbol{\mathcal{C}}$, $\boldsymbol{\mathcal{B}}_{ego}$ and $\boldsymbol{\mathcal{B}}_{surr}$ and SVs' reactions $\boldsymbol{U}_{surr}$ in the future horizon $N$ as illustrated in Eq. (25), where $N$ denotes the future horizon length; $\theta$ denotes the parameters of the proposed socially-aware encoder-decoder.

*D. Socially-aware MPC planner*

In this section, a socially-aware MPC planner is designed. It accounts for the social interaction mechanism by considering the explicit social interaction dynamics as the constraint under a leader-follower game framework [34]. This design enables the planner to generate manifold, human-like behaviors when interacting with surrounding traffic while mitigating the potential safety risks.

**Definition 1**: For solving the socially-aware MPC planner in a rolling horizon fashion, some vector and matrix variables in the future horizon dimension shall be defined as follows. $N$ also denotes the discrete future horizon length.

$$\boldsymbol{\mathcal{X}} = \left(\boldsymbol{X}_t, \boldsymbol{X}_{t+1}, \dots, \boldsymbol{X}_{t+N}\right)^{\mathrm{T}} \tag{26}$$

$$\boldsymbol{\mathcal{U}}_{ego} = \left(\boldsymbol{U}_{ego,t}, \boldsymbol{U}_{ego,t+1}, \dots, \boldsymbol{U}_{ego,t+N-1}\right)^{\mathrm{T}} \tag{27}$$

$$\boldsymbol{\mathcal{U}}_{surr} = \left(\boldsymbol{U}_{surr,t}, \boldsymbol{U}_{surr,t+1}, \dots, \boldsymbol{U}_{surr,t+N-1}\right)^{\mathrm{T}} \tag{28}$$

$$\overline{\boldsymbol{\mathcal{A}}} = \begin{bmatrix} \mathbf{0} & \mathbf{0} & \cdots & \mathbf{0} & \mathbf{0} \\ \boldsymbol{\mathcal{A}}_t & \mathbf{0} & \cdots & \mathbf{0} & \mathbf{0} \\ \mathbf{0} & \ddots & \ddots & \vdots & \vdots \\ \vdots & \mathbf{0} & \ddots & \mathbf{0} & \vdots \\ \mathbf{0} & \mathbf{0} & \cdots & \boldsymbol{\mathcal{A}}_{t+N-1} & \mathbf{0} \end{bmatrix} \tag{29}$$

$$\overline{\boldsymbol{\mathcal{B}}}_{ego} = \begin{bmatrix} \mathbf{0} & \mathbf{0} & \cdots & \mathbf{0} \\ \boldsymbol{\mathcal{B}}_{ego,t} & \mathbf{0} & \cdots & \mathbf{0} \\ \mathbf{0} & \ddots & \ddots & \vdots \\ \vdots & \mathbf{0} & \ddots & \mathbf{0} \\ \mathbf{0} & \mathbf{0} & \cdots & \boldsymbol{\mathcal{B}}_{ego,t+N-1} \end{bmatrix} \tag{30}$$

$$\boldsymbol{C}_{veh_1,veh_2,t}, \boldsymbol{C}_{veh_1,veh_2,t+1}, \dots, \boldsymbol{C}_{veh_1,veh_2,t+N-1} = TransformerDecoderLayer\left(\mathbb{F}_{veh_1\to veh_2,t}\right) \tag{23}$$

$$veh_1, veh_2 \in \{EAV, FV, RV, LFV, LRV, RFV, RRV\}, veh_1 \neq veh_2$$

$$\boldsymbol{B}_{veh_1,veh_2,t}, \boldsymbol{B}_{veh_1,veh_2,t+1}, \dots, \boldsymbol{B}_{veh_1,veh_2,t+N-1} = TransformerDecoderLayer\left(\mathbb{F}_{veh_1\to veh_2,t}\right) \tag{24}$$

$$veh_1, veh_2 \in \{EAV, FV, RV, LFV, LRV, RFV, RRV\}, veh_1 \neq veh_2$$

$$\begin{bmatrix} \boldsymbol{\mathcal{C}}_t, \boldsymbol{\mathcal{C}}_{t+1}, \dots, \boldsymbol{\mathcal{C}}_{t+N-1} \\ \boldsymbol{\mathcal{B}}_{ego,t}, \boldsymbol{\mathcal{B}}_{ego,t+1}, \dots, \boldsymbol{\mathcal{B}}_{ego,t+N-1} \\ \boldsymbol{\mathcal{B}}_{surr,t}, \boldsymbol{\mathcal{B}}_{surr,t+1}, \dots, \boldsymbol{\mathcal{B}}_{surr,t+N-1} \\ \boldsymbol{U}_{surr,t}, \boldsymbol{U}_{surr,t+1}, \dots, \boldsymbol{U}_{surr,t+N-1} \end{bmatrix} = \mathrm{SociallyAwareEncoderDecoder}_{\theta}\begin{pmatrix} \boldsymbol{Traj}_{ego,t}, \dots, \boldsymbol{Traj}_{surr_n,t}, \\ \boldsymbol{Map}_{veh,t}, \dots, \boldsymbol{Map}_{surr_n,t}, \\ \boldsymbol{U}_{ego,t}, \boldsymbol{U}_{ego,t+1}, \dots, \\ \boldsymbol{U}_{ego,t+N-1} \end{pmatrix} \tag{25}$$

$$\overline{\mathcal{B}}_{surr}=\begin{bmatrix}\mathbf{0} & \mathbf{0} & \cdots & \mathbf{0}\\ \mathcal{B}_{surr,t} & \mathbf{0} & \cdots & \mathbf{0}\\ \mathbf{0} & \ddots & \ddots & \vdots\\ \vdots & \mathbf{0} & \ddots & \mathbf{0}\\ \mathbf{0} & \mathbf{0} & \cdots & \mathcal{B}_{surr,t+N-1}\end{bmatrix} \tag{31}$$

$$\overline{\mathcal{C}}=\begin{bmatrix}\mathbf{0} & \mathbf{0} & \cdots & \mathbf{0} & \mathbf{0} & \mathbf{0}\\ \mathcal{C}_t & \mathbf{0} & \cdots & \mathbf{0} & \mathbf{0} & \mathbf{0}\\ -\mathcal{C}_t & \mathcal{C}_{t+1} & \ddots & \mathbf{0} & \vdots & \vdots\\ \vdots & \mathbf{0} & \ddots & \mathcal{C}_{t+N-2} & \mathbf{0} & \vdots\\ \mathbf{0} & \mathbf{0} & \cdots & -\mathcal{C}_{t+N-2} & \mathcal{C}_{t+N-1} & \mathbf{0}\end{bmatrix} \tag{32}$$

$$\overline{\mathcal{D}}=(\boldsymbol{X}_t,\mathbf{0},\dots,\mathbf{0})^T \tag{33}$$

$$\boldsymbol{Q}=\operatorname{diag}(\theta_1,\theta_2,\theta_3,\theta_4,\underbrace{0,\dots,0}_{4n}) \tag{34}$$

$$\boldsymbol{R}=\operatorname{diag}(\theta_5,\theta_6,\underbrace{0,\dots,0}_{2n}) \tag{35}$$

$$\boldsymbol{X}_{des}=(s_{des},v_{des},y_{des},\psi_{des},\underbrace{0,\dots,0}_{4n}) \tag{36}$$

where $\mathcal{X}$, $\mathcal{U}_{ego}$, and $\mathcal{U}_{surr}$ denote the state vector, control inputs of EAV, and control inputs of SVs in the future planning horizon respectively; $\mathcal{U}_{surr}$ can be seen as the **SV's reactions** produced by the socially-aware encoder-decoder (Section II.C); $\overline{\mathcal{A}}$, $\overline{\mathcal{C}}$, $\overline{\mathcal{B}}_{ego}$, and $\overline{\mathcal{B}}_{surr}$ are the state matrix and control matrices in the future planning horizon respectively. $\boldsymbol{Q}$ and $\boldsymbol{R}$ are coefficient matrices; $\theta_1,\theta_2,\theta_3,\theta_4,\theta_5,\theta_6$ denote the weighting parameters used in the cost function; $s_{des},v_{des},y_{des},\psi_{des}$ denote the desired driving states of the EAV (position, speed, and heading angle). The weighting parameters are configured by previous experience and hand-crafted tuning. The values are 10, 10, 10, 100, 10, and 50.

*1) Leader-follower game framework*

This paper provides socially-aware and reliable motions based on a leader-follower game framework. As a form of non-cooperative game, the leader acts first, and the follower responds based on observations. For this paper, the EAV is modeled as the leader and the SVs as followers, rather than implying a fixed role in real-world driving. The EAV anticipates the rational reactions of SVs and proactively optimizes its motions. SVs are assumed to react to EAV's motions. The socially-aware encoder-decoder anticipates their optimal reactions. The strategic social interactions between the EAV and SVs are embedded into a constrained optimization problem, expressed as in TABLE 3.

In TABLE 3, $J_{EAV}(\mathcal{X},\mathcal{U}_{ego})$ represents the EAV's driving cost; $\mathcal{G}(\mathcal{X},\mathcal{U}_{ego},\overline{\mathcal{A}},\overline{\mathcal{B}}_{ego},\overline{\mathcal{B}}_{surr},\overline{\mathcal{C}},\mathcal{U}_{surr})=\mathbf{0}$ denotes the explicit social interaction dynamics. By embedding SVs' reactions $\mathcal{U}_{surr}$ and data-driven social interaction matrices $\overline{\mathcal{C}}$, $\overline{\mathcal{B}}_{ego}$, and $\overline{\mathcal{B}}_{surr}$ into the constraint set, this formulation maintains a tractable leader-follower game optimization structure in which **EAV optimizes its motions with consideration of SVs reactions and social interaction effects.** The implementation details of the leader-follower game framework are illustrated in Figure 6. The socially-aware encoder-decoder provides anticipated SVs' reactions and other data-driven social interaction matrices (Section II.B) to formulate explicit social interaction dynamics constraints.

TABLE 3 The constrained leader-follower game optimization problem.

| **The constrained leader-follower game optimization problem modeling** | |
|---|---|
| 1: | **Cost function:** $min\, J_{EAV}\,(\mathcal{X},\mathcal{U}_{ego})$ |
| 2: | **Constraints:** |
| 3: | Socially-aware constraint: $\mathcal{G}(\mathcal{X},\mathcal{U}_{ego},\overline{\mathcal{A}},\overline{\mathcal{B}}_{ego},\overline{\mathcal{B}}_{surr},\overline{\mathcal{C}},\mathcal{U}_{surr})=\mathbf{0}$ |
| 4: | $\overline{\mathcal{C}},\overline{\mathcal{B}}_{ego},\overline{\mathcal{B}}_{surr},\mathcal{U}_{surr}=\text{SociallyAwareEncoderDecoder}_\theta\begin{pmatrix}\boldsymbol{Traj}_{ego,t},\dots,\boldsymbol{Traj}_{surr_n,t},\\ \boldsymbol{Map}_{ego,t},\dots,\boldsymbol{Map}_{surr_n,t},\mathcal{U}_{ego}\end{pmatrix}$ |
| 5: | Other constraints (collision avoidance, speed limit, …) |
| 6: | **Solving:** any quadratic programming algorithm (We select the Gurobi solver) |

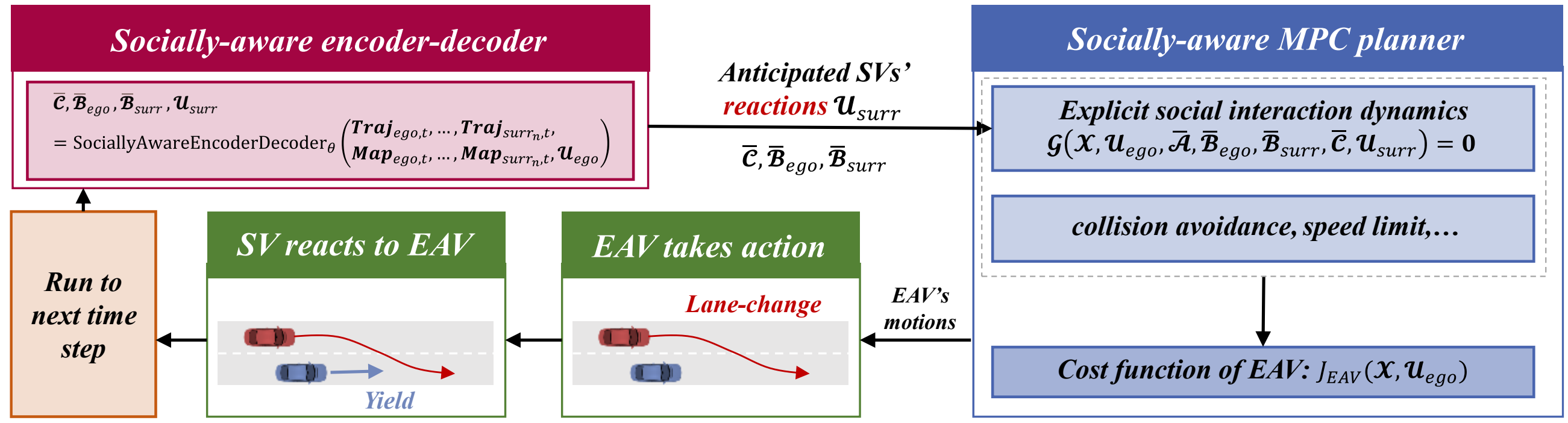

Figure 6 The leader-follower game framework

$$\begin{aligned}J_{EAV}=\frac{1}{2}\sum_{t=0}^{N-1}\Big[&\underbrace{\theta_1\left(s_{ego,t}-s_{des}\right)^2}_{\text{longitudinal target}}+\underbrace{\theta_2\left(v_{ego,t}-v_{des}\right)^2}_{\text{driving efficiency}}+\underbrace{\theta_3 a_{ego,t}^2}_{\text{ride comfort}}\\ &+\underbrace{\theta_4\left(y_{ego,t}-y_{des}\right)^2}_{\text{lane-changing request}}+\underbrace{\theta_5\psi_{ego,t}^2}_{\text{heading-changing smoothness}}+\underbrace{\theta_6\delta_{f_{ego,t}}^{\;2}}_{\text{steering smoothness}}\Big]\end{aligned} \tag{37}$$

The socially-aware MPC planner optimizes EAV's motions considering SVs' reactions. EAV, as the leader, takes action first. Then, SV, as the follower, conducts a reaction that completes the interactive cycle in the leader-follower game context. Finally, the whole process runs to the next time step.

*2) Cost function*

This paper designs a quadratic form of the cost function for the EAV as shown in Eq. (37). It reflects a combination of longitudinal and lateral driving goals. Taking a lane-changing maneuver as an example, the first three terms drive the vehicle towards the target longitudinal position and speed while ensuring ride comfort. The fourth term drives the vehicle towards the target lane. The latter two terms confirm the heading-changing and steering smoothness. Then the cost function is converted to matrix form:

$$min\, J_{EAV} = \frac{1}{2}\begin{bmatrix}\boldsymbol{\mathcal{X}}^T & \boldsymbol{\mathcal{U}}_{ego}{}^T\end{bmatrix}\boldsymbol{\mathcal{P}}\begin{bmatrix}\boldsymbol{\mathcal{X}}\\ \boldsymbol{\mathcal{U}}_{ego}\end{bmatrix} + \boldsymbol{q}^T\begin{bmatrix}\boldsymbol{\mathcal{X}}\\ \boldsymbol{\mathcal{U}}_{ego}\end{bmatrix} \tag{38}$$

where,

$$\boldsymbol{\mathcal{P}} = diag\left(\underbrace{\boldsymbol{Q},\dots,\boldsymbol{Q}}_{N}, \boldsymbol{0}_{5\times5}, \underbrace{\boldsymbol{R},\dots,\boldsymbol{R}}_{N}\right) \tag{39}$$

$$\boldsymbol{q} = \left(\underbrace{-\boldsymbol{Q}\boldsymbol{X}_{des},\dots,-\boldsymbol{Q}\boldsymbol{X}_{des}}_{N}, 0,\dots,0\right)^T \tag{40}$$

*3) Constraints*

**Socially-aware constraint**: Under the leader-follower game framework, this paper accounts for the social interaction mechanism by considering the explicit social interaction dynamics $\boldsymbol{\mathcal{G}}(\boldsymbol{\mathcal{X}}, \boldsymbol{\mathcal{U}}_{ego}, \bar{\boldsymbol{\mathcal{A}}}, \bar{\boldsymbol{\mathcal{B}}}_{ego}, \bar{\boldsymbol{\mathcal{B}}}_{surr}, \bar{\boldsymbol{\mathcal{C}}}, \boldsymbol{\mathcal{U}}_{surr}) = \boldsymbol{0}$ as the constraint. This constraint incorporates the data-driven social interaction matrices and the reactions of SVs into the motion planning of EAV. This constraint is transformed to the dimension of the future optimization horizon:

$$\boldsymbol{\mathcal{X}} = (\bar{\boldsymbol{\mathcal{A}}} + \bar{\boldsymbol{\mathcal{C}}})\boldsymbol{\mathcal{X}} + \bar{\boldsymbol{\mathcal{B}}}_{ego}\boldsymbol{\mathcal{U}}_{ego} + \bar{\boldsymbol{\mathcal{B}}}_{surr}\boldsymbol{\mathcal{U}}_{surr} + \bar{\boldsymbol{\mathcal{D}}} \tag{41}$$

then Eq. (41) is converted to align with the decision variables in Eq. (38).

---

**Algorithm 1**: Adding pairwise collision avoidance constraints between EAV and SVs

1: **Require:** large enough positive number M; longitudinal safety distance reference $s_{ref}$; lateral safety distance reference $y_{ref}$; planning horizon $N$; EAV and SVs $surr_1$ to $surr_n$
2: **for** $j \leftarrow 1$ to $N$ **do:**
3:     **for** $SV \leftarrow surr_1$ to $surr_n$ **do:**
4:         Initialize a binary variable $c_{SV,j} \in \{0,1\}$
5:         Adding Big-M constraints between EAV and current SV:
        $|s_{SV,j} - s_{ego,j}| \geq s_{ref} - \mathrm{M}\cdot c_{SV,j}$
        $|y_{SV,j} - y_{ego,j}| \geq y_{ref} - \mathrm{M}\cdot(1 - c_{SV,j})$
6:     **end for**
7: **end for**

---

$$s.t.\begin{bmatrix}\bar{\boldsymbol{\mathcal{A}}} + \bar{\boldsymbol{\mathcal{C}}} - \boldsymbol{I} & \bar{\boldsymbol{\mathcal{B}}}_{ego}\end{bmatrix}\begin{bmatrix}\boldsymbol{\mathcal{X}}\\ \boldsymbol{\mathcal{U}}_{ego}\end{bmatrix} + \bar{\boldsymbol{\mathcal{B}}}_{surr}\boldsymbol{\mathcal{U}}_{surr} + \bar{\boldsymbol{\mathcal{D}}} = \boldsymbol{0} \tag{42}$$

the socially-aware encoder-decoder provides anticipated SVs' reactions and other data-driven social interaction matrices. It can be transformed from Eq. (25) as illustrated in Eq. (43).

**Collision avoidance constraints**: This paper also accounts for collision avoidance using Big-M Method [35] as illustrated in Algorithm 1. These constraints ensure the EAV maintains either a longitudinal or lateral safe distance from SVs.

**Other constraints**: This paper also considers other constraints, such as speed range, acceleration range, and heading range.

*D. Learning & planning framework*

This paper trains the socially-aware encoder-decoder proposed in MPCFormer using a natural driving dataset. During the forward pass, the proposed approach plans EAV

$$\bar{\boldsymbol{\mathcal{C}}}, \bar{\boldsymbol{\mathcal{B}}}_{ego}, \bar{\boldsymbol{\mathcal{B}}}_{surr}, \boldsymbol{\mathcal{U}}_{surr} = \mathrm{SociallyAwareEncoderDecoder}_{\theta}\begin{pmatrix}\boldsymbol{Traj}_{ego,t}, \dots, \boldsymbol{Traj}_{surr_n,t},\\ \boldsymbol{Map}_{ego,t}, \dots, \boldsymbol{Map}_{surr_n,t}, \boldsymbol{\mathcal{U}}_{ego}\end{pmatrix} \tag{43}$$

---

**Algorithm 2**: Learning & planning

**Require:** SociallyAwareEncoderDecoder$_{\theta}$, planning steps $N_p$, training steps $T_s$, dataset $\mathcal{DS}$
1: **for** $j \leftarrow 1$ to $N_t$ **do:**
2:     Randomly sample a batch of data from dataset $\mathcal{DS}$
3:     Plan EAV's motions $\boldsymbol{\mathcal{U}}_{ego}$ to follow the ground-truth ego states
4:     Predict $\bar{\boldsymbol{\mathcal{C}}}$, $\bar{\boldsymbol{\mathcal{B}}}_{ego}$, $\bar{\boldsymbol{\mathcal{B}}}_{surr}$ and $\boldsymbol{\mathcal{U}}_{surr}$ through SociallyAwareEncoderDecoder$_{\theta}$
5:     Calculate system states $\boldsymbol{\mathcal{X}}$ using explicit social interaction dynamics
6:     Calculate total loss $\mathcal{L}$ according to Eq. (46)
7:     Back-propagate loss and calculate gradients with respect to parameters $\theta$
8:     Update InteractionAwareEncoderDecoder$_{\theta}$
9: **end for**
10: Initialize the interactive simulation environment
11: **for** $t \leftarrow 1$ to $N_p$ **do:**
12:     Predict $\bar{\boldsymbol{\mathcal{C}}}$, $\bar{\boldsymbol{\mathcal{B}}}_{ego}$, $\bar{\boldsymbol{\mathcal{B}}}_{surr}$ and $\boldsymbol{\mathcal{U}}_{surr}$ using trained SociallyAwareEncoderDecoder$_{\theta}$
13:     Plan EV's motion $\boldsymbol{\mathcal{U}}_{ego}$ using the proposed Socially-aware MPC planner (solved by Gurobi)
14:     Execute the first step of the planned motion for EAV
15: **end for**

motions to track the ground-truth ego states while predicting SVs' states via the explicit social interaction dynamics. In the backward pass, this approach evaluates the loss function on the predicted and planned states. The loss function includes vehicle state loss and GMM loss. Vehicle state loss is the smooth L1 error between the ground truth vehicle states and the predicted and planned states:

$$\mathcal{L}_{vehicles} = SmoothL_1\left(\boldsymbol{\mathcal{X}} - \boldsymbol{\mathcal{X}}_{groundtruth}\right) \tag{44}$$

The GMM loss is to minimize the variances of the selected $\boldsymbol{U}_{surr}$ modality and the negative logarithm of the selected modality probability:

$$\mathcal{L}_{GMM} = \log\left(\hat{\sigma}_{\boldsymbol{U}_{surr}}\right) - \log\left(\hat{p}\right) \tag{45}$$

Hence, the loss function in the learning stage can be expressed as follows, in which $\lambda_1$ and $\lambda_2$ are weighting factors:

$$\mathcal{L} = \lambda_1 \mathcal{L}_{vehicles} + \lambda_2 \mathcal{L}_{GMM} \tag{46}$$

In the planning stage, this paper conducts motion planning for accomplishing EAV's driving tasks in an interactive simulation environment. The motion planning is implemented in a rolling horizon fashion. Detailed learning & planning information is depicted in Algorithm 2.

## III. EVALUATION

The evaluation of the proposed MPCFormer is divided into two parts: i) open-looped prediction performance validation; ii) close-looped planning performance validation. The open-looped prediction is to validate the SVs' trajectory prediction performance of the proposed approach. The close-looped planning performance is to validate the EAV's planning performance.

### A. *Open-looped prediction performance validation*

The open-looped prediction performance validation aims to answer the following questions: i) How is the proposed MPCFormer's prediction capability compared to existing approaches? ii) Can the proposed MPCFormer actually enable explainable, socially-aware predictions?

#### 1) *Dataset preparation*

This paper adopts the Next Generation Simulation (NGSIM) trajectory dataset [36], covering the whole 640-meter section of US Highway 101 with five main lanes and an auxiliary lane between an on-ramp and an off-ramp, as shown in Figure 7. This paper segments the whole dataset into frames using the sliding-window strategy to improve data efficiency [37]. Each frame is treated as an independent prediction instance with fixed history and prediction horizons, containing 9-s object tracks at 10 Hz, which consist of a 4-s history observation horizon and a 5-s prediction horizon.

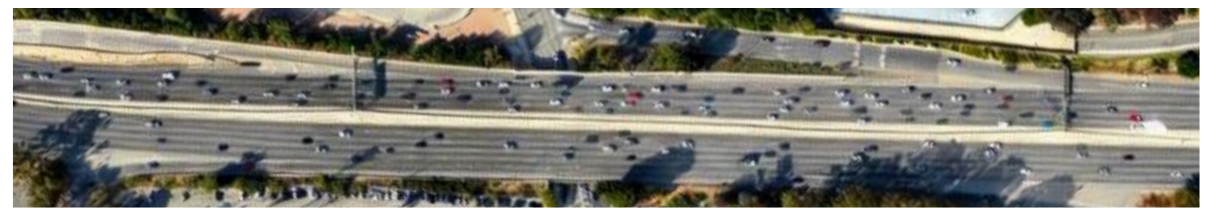

Figure 7 Data record area in US Highway 101.

#### 2) *Measures of effectiveness*

The following Measures of Effectiveness (MOE) are utilized for open-looped prediction performance validation:

- Average Displacement Error (ADE) and Final Displacement Error (FDE) are used to measure the prediction advantage of the proposed approach. ADE is defined as the average L2 norm distance between each point of the ground-truth trajectories and predicted trajectories. FDE is defined as the L2 norm distance between the final point of the ground-truth trajectories and predicted trajectories.

$$ADE = \frac{\sum_{n \in N} \sum_{t \in N_p} ||\hat{p}_t^n - p_t^n||_2}{N \times N_p} \tag{47}$$

$$FDE = \frac{\sum_{n \in N} ||\hat{p}_{N_p}^n - p_{N_p}^n||_2}{N} \tag{48}$$

where $\hat{p}_t^n$ denotes the predicted position of vehicle $n$ at time $t$; $p_t^n$ denotes the ground-truth position of vehicle $n$ at time $t$.

- The predicted future trajectories of SVs are used for prediction qualitative analysis, and prediction explainability validation.
- The Frobenius norms of data-driven social interaction matrices are used to explicitly explain the social interaction strength between vehicles. It is defined as the square root of the sum of the absolute squares of all elements in a matrix. The higher value of the Frobenius norm indicates the higher social interaction strength between vehicles.

#### 3) *Implementation details*

Based on the vehicle track ID, this paper selects 70% of processed data for training, 10% for validation, and 20% for testing. The training epoch is 25, and the learning rate is $1 \times 10^{-3}$. The model is trained on an NVIDIA RTX 4080 GPU with a batch size of 256. This paper trained the model five times at different random seeds.

#### 4) *Baselines*

The following baseline approaches are selected because of great academic reputation: **S-GAN** [38]: a multi-modal human trajectory prediction GAN trained with a variety loss to encourage diversity; **STGAT** [39]: a social spatial-temporal graph convolutional neural network for human trajectory prediction; **PECNet** [42]: a VAE based model with goal conditioning predictions; Others are state-of-the-art methods: **STM** [40]: a spatial-aware transformer for vehicle trajectory prediction; **VCIFormer** [41]: a velocity-aware complementary interaction transformer; **C2F-TP** [43]: a coarse-to-fine diffusion transformer framework for uncertainty-aware vehicle trajectory prediction; **SSTT** [44]: an interaction-aware trajectory prediction method based on sparse spatial-temporal transformer; MPIDL [45]: a mixed-granularity physics-informed deep learning framework.

#### 5) *Quantification results*

This subsection aims to answer the first question. The open-looped prediction performance is quantified through prediction errors as shown in TABLE 4. Results demonstrate that MPCFormer achieves robust short- and long-term prediction performance, outperforming most approaches across all horizons with the lowest ADEs of 0.12, 0.28, 0.48, 0.69, and 0.86 m at 1–5 s, respectively. Although SSTT achieves the same ADE@1s, its prediction errors increase beyond 1 s. MPCFormer also maintains a competitive FDE of 2.20 m. While PECNet achieves the lowest FDE due to its endpoint-conditioned prediction design, MPCFormer provides more balanced and consistently accurate performance across different metrics and prediction horizons.

#### 6) *Qualitative results*

This subsection also aims to answer the first question. Figure 8 shows some representative interactive scenarios, demonstrating that the proposed approach has great open-looped prediction performance. Figure 8 (a) and (b) display the longitudinal-only scenarios, which are EAV following in a slower lane and faster lane, respectively. These two scenarios highlight the ability of the proposed approach to accurately predict trajectories in both fast and slow lanes at the same time. Figure 8 (c) and (d) are two lane-changing scenarios, which are EAV being cut-in and EAV's front SV lane-changing. They demonstrate that the proposed approach shows its superior performance in predicting SV's lateral maneuver.

TABLE 4 Prediction error comparison in different time horizons.

| Approach | Prediction error (m) | | | | | |
|---|---|---|---|---|---|---|
| | **ADE@1s** | **ADE@2s** | **ADE@3s** | **ADE@4s** | **ADE@5s** | **FDE** |
| S-GAN | 0.19 | 0.34 | 0.57 | 0.83 | 1.21 | 2.55 |
| STGAT | 0.28 | 0.58 | 0.95 | 1.39 | 1.88 | 4.35 |
| PECNet | 0.28 | 0.50 | 0.68 | 0.82 | 0.88 | **0.43** |
| STM | 0.23 | 0.54 | 1.01 | 1.63 | 2.08 | 2.88 |
| VCIFormer | - | - | - | - | 0.95 | 2.22 |
| MPIDL | 0.16 | 0.46 | 0.71 | 0.96 | 1.33 | 2.84 |
| C2F-TP | 0.20 | 0.47 | 0.78 | 1.08 | 1.45 | 1.36 |
| SSTT | **0.12** | 0.48 | 0.89 | 1.34 | 1.91 | 2.18 |
| **MPCFormer** | **0.12** | **0.28** | **0.48** | **0.69** | **0.86** | 2.20 |

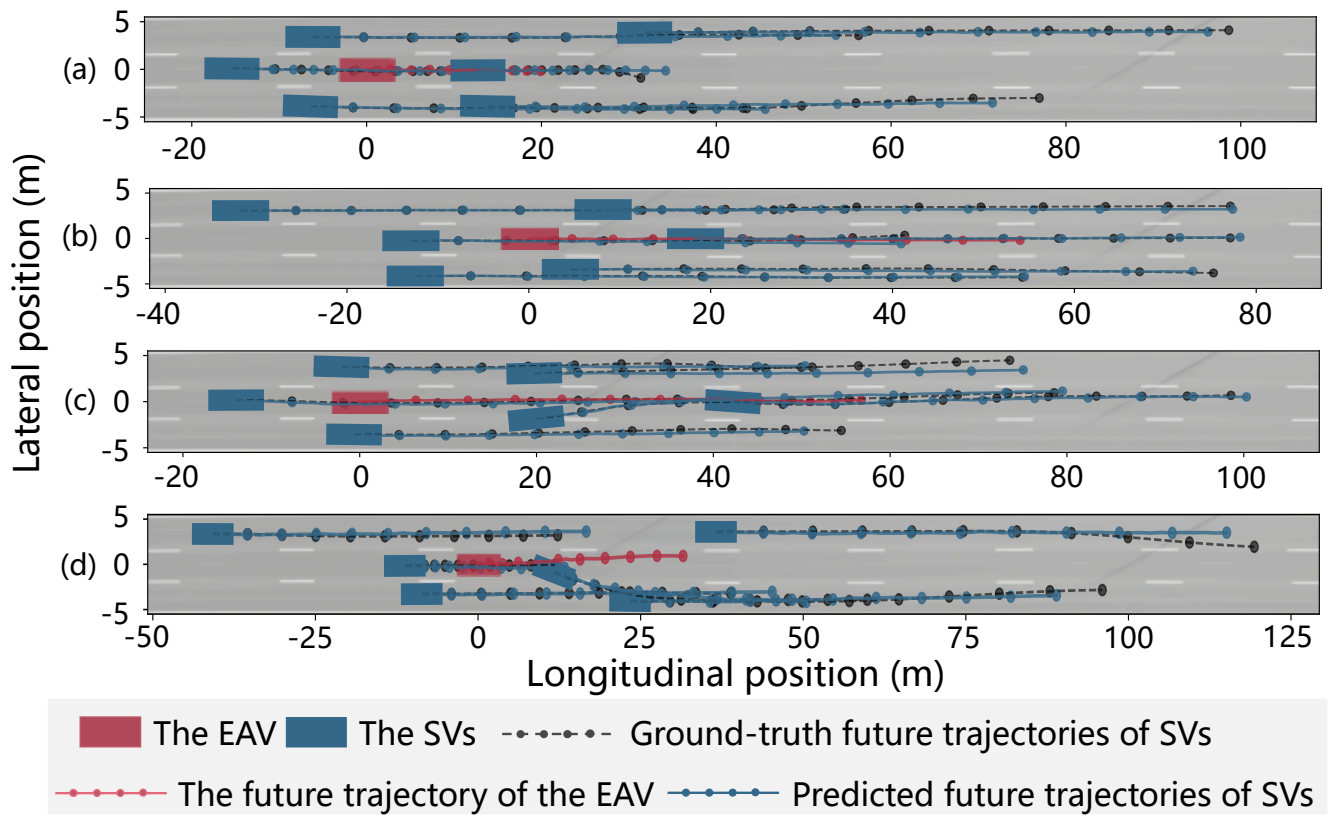

Figure 8 Qualitative results of the proposed approach. (a) EAV following in slower lane; (b) EAV following in faster lane; (c) EAV being cut-in; (d) EAV' front SV lane-changing (The prediction horizon is 5 seconds, and the mark points denote the vehicles' position every 0.5 seconds.

#### 7) *Socially-aware predictions with explainability*

This subsection aims to answer the second question. Figure 9 and TABLE 5 confirm that the proposed MPCFormer actually enables socially-aware predictions with explainability through a randomly selected case. Figure 9 (a) displays the original prediction, which means predicted future trajectories of SVs under the ground-truth future trajectories of the EAV. In Figure 9 (b), this paper modifies the ground-truth future trajectory of the EAV to perform an accelerating and left lane-changing maneuver and analyzes how this change affects the predicted future trajectories of the SVs. TABLE 5 shows that the modified trajectory of the EAV demonstrates 5.28 meters and 3.69 meters more than the original trajectory in longitudinal and lateral coordinates. It can be seen that the SVs exhibit relatively steady and lane-consistent predicted motions in Figure 9 (a). In contrast, Figure 9 (b) shows that the Front Vehicle (FV) is strongly influenced by the EAV's modified motion, showing anticipatory acceleration in response to the EV's upcoming maneuver. Compared with the original prediction, the FV travels 4.11 m farther with a 1.32 m/s higher speed. In contrast, the Left Front Vehicle (LFV) shows only slight acceleration, likely because it is farther ahead and the lane change occurs behind it. These responses demonstrate the model's ability to capture socially aware behaviors. Other SVs remain largely unaffected; for example, the Rear Right Vehicle (RRV) exhibits a longitudinal trajectory difference of less than 0.3 m, consistent with the limited influence of the EV's maneuver on vehicles directly behind it.

Figure 9 (c) and TABLE 6 validate the explainability of the prediction with the data-driven social interaction matrices defined in Section 3.2.2. For instance, $\overline{\boldsymbol{B}_{ego,FV}}$ explicitly represents the time-averaged social interaction effect from the EAV to the FV. These matrices vary across vehicle pairs, indicating that the model captures distinct and directional social interaction patterns between different vehicles. Moreover, TABLE 6 shows that the EAV–FV interaction has the highest norm value (0.70), indicating the strongest influence on the FV. In contrast, interactions with rear-side vehicles, such as RV and RRV, are much weaker (0.26 and 0.27). This explicit interaction dynamics modeling enhances explainability by revealing how vehicles influence each other over time.

### B. *Close-looped planning performance validation*

The close-looped planning performance validation aims to answer the following questions: i) How is the proposed MPCFormer's planning capability compared to existing approaches? ii) Can the proposed MPCFormer enable explainable socially-aware planning? iii) Can the proposed MPCFormer enhance driving efficiency without being unsafe?

#### 1) *test scenario*

As illustrated in Figure 10, the test scenario takes place on a section of interweaving traffic on US Highway 101. The goal of EAV is to navigate and successfully complete an off-ramp maneuver. The start position and the target lane of EAV are marked in Figure 10. The EAV must complete three consecutive lane changes within a short distance, leading to intense interactions, making this a challenging scenario. Six test cases are designed as shown in TABLE 8. Sensitivity analysis is conducted in terms of traffic congestion levels (v/c ratio of 0.4, 0.6, and 0.8) and driving styles (aggressive and normal).

#### 2) *Simulation platform for close-looped driving*

The planning simulation platform is Vissim [46], which is a highly realistic traffic and vehicle dynamics simulation software. The reason why this paper builds a new simulation environment instead of replaying NGSIM dataset directly like many current researches [6] is that:

- If replaying NGSIM data directly, SVs cannot respond to the EAV's planned motion, which is called "non-reactive" vehicles. By contrast, Vissim has highly developed reaction

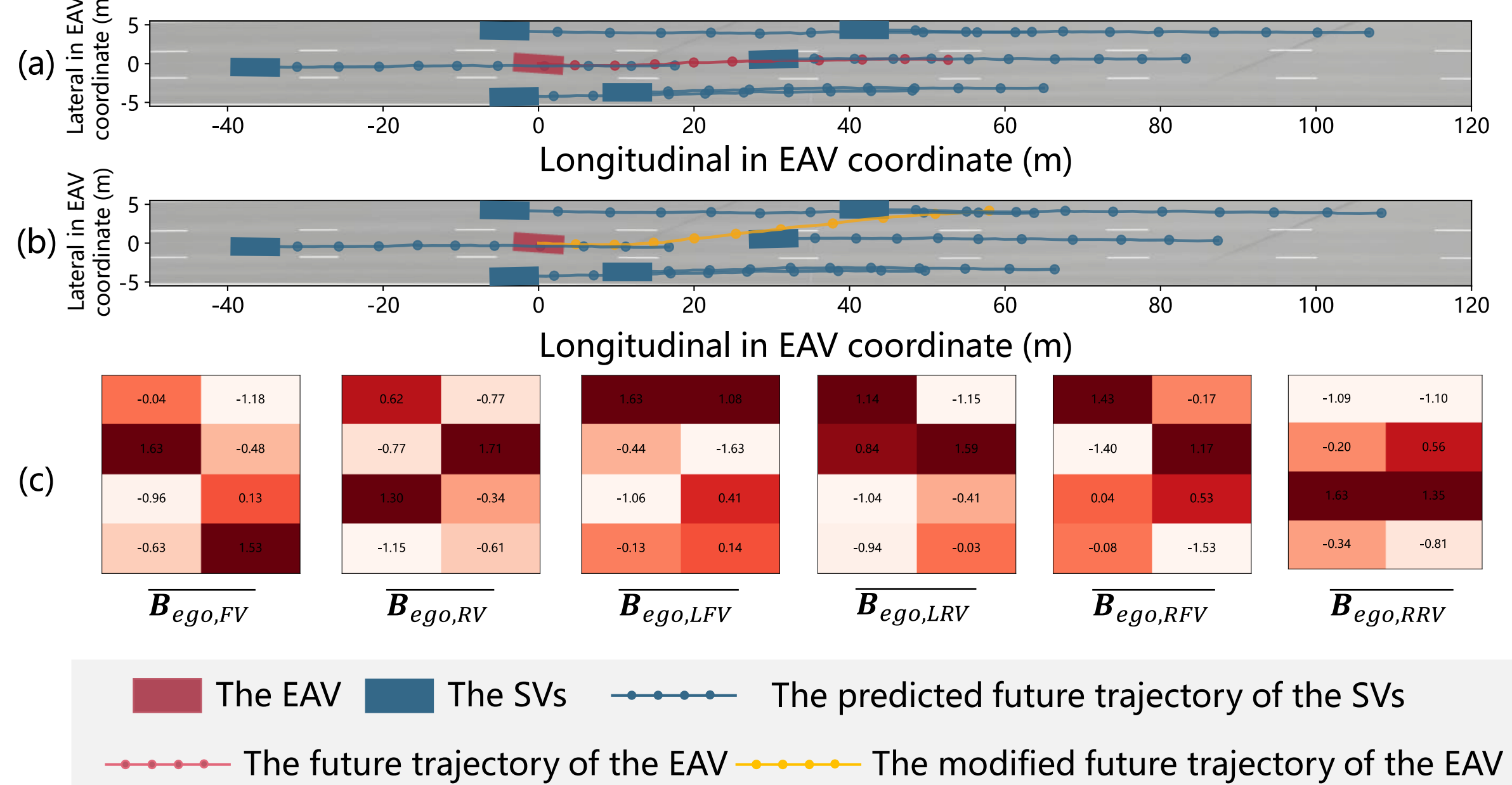

Figure 9 Validation for socially-aware predictions with explainability. (a) original prediction; (b) prediction after modifying EAV's future trajectory; (c) the time-averaged data-driven social interaction matrices defined in 3.2.2. (Additional definition: FV-Front Vehicle, RV-Rear Vehicle, LFV-Left Front Vehicle, LRV-Left Rear Vehicle, RFV-Right Front Vehicle, and RRV-Right Rear Vehicle)

TABLE 5 Trajectory difference (predicted trajectories after modifying the EAV's future trajectory Figure 9 (b) minus the original predicted trajectories Figure 9 (a)).

| **Trajectory difference** | **EAV** | **FV** | **RV** | **LFV** | **LRV** | **RFV** | **RRV** |
|---|---|---|---|---|---|---|---|
| Longitudinal (m) | 5.28 | 4.11 | -0.77 | 1.42 | 1.62 | 1.58 | 0.27 |
| Lateral (m) | 3.69 | -0.36 | -0.25 | -0.22 | -0.08 | -0.12 | -0.20 |
| Speed (m/s) | 3.11 | 1.32 | -0.10 | 0.60 | 0.71 | 0.63 | 0.25 |

TABLE 6 The Frobenius norm value of data-driven social interaction matrices in Figure 9 (c)

| | $\overline{\boldsymbol{B}_{ego,FV}}$ | $\overline{\boldsymbol{B}_{ego,RV}}$ | $\overline{\boldsymbol{B}_{ego,LFV}}$ | $\overline{\boldsymbol{B}_{ego,LRV}}$ | $\overline{\boldsymbol{B}_{ego,RFV}}$ | $\overline{\boldsymbol{B}_{ego,RRV}}$ |
|---|---|---|---|---|---|---|
| **Frobenius norm value** | 0.70 | 0.26 | 0.35 | 0.36 | 0.36 | 0.27 |

TABLE 7 Calibration results of the simulation platform compared to NGSIM

| **Comparison** | **NGSIM** | **The calibrated simulation platform** | **Error** |
|---|---|---|---|
| Main-lane entrance (veh/h) | 3240 | 3247±54 | 0.21% |
| On-ramp entrance (veh/h) | 503 | 509±18 | 1.19% |
| Main-lane left (veh/h) | 3283 | 3481±70 | 6.03% |
| Off-ramp left (veh/h) | 271 | 274±8 | 1.11% |
| Traffic average speed (km/h) | 34.23 | 35.18±0.63 | 2.78% |
| Traffic lane-changing times | 5198 | 5107±173 | 1.75% |

models for SVs. Hence, it can portray realistic social interactions between EAV and SVs.

- Vissim can support flexible sensitivity analysis, but replaying NGSIM data cannot.

By calibrating traffic parameters, this paper sets up a simulation environment similar to NGSIM dataset. The calibration results are shown in TABLE 7.

*3) Planners for comparison*

Three types of socially-aware planners are considered:

- **The PAS planner:** A traditional **PAssively Socially-aware (PAS)** approach designed based on the Intelligent Driver Model (IDM) [47] and Minimizing Overall Braking Induced by Lane-change (MOBIL) model [48].

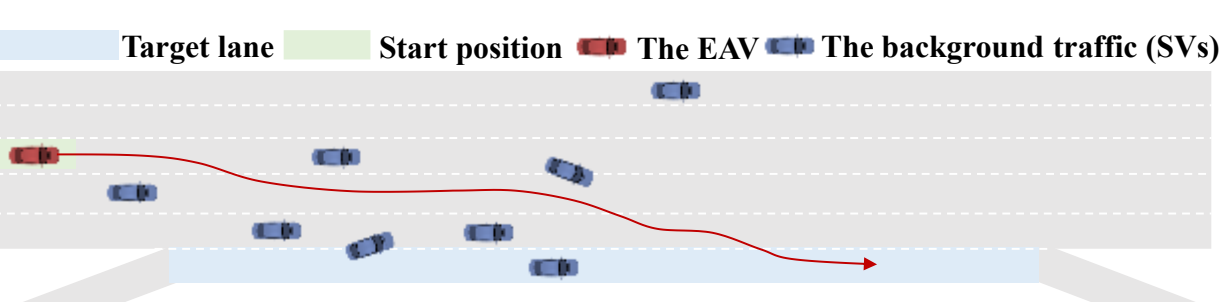

Figure 10 The test scenario: EAV off-ramp in US Highway 101.

TABLE 8 Case definitions

| | v/c = 0.4 | v/c =0.6 (NGSIM) | v/c = 0.8 |
|---|---|---|---|
| Aggressive (NGSIM) | Case1 | **Case2** | Case3 |
| Normal | Case4 | Case5 | Case6 |

- **The NS planner:** A Proximal Policy Optimization (PPO) reinforcement learning approach for lane-changing [17]. This approach is a classic **Neutrally Socially-aware (NS)** approach and has gained high academic impact since its publication. The reward function and parameter settings remain the same with the original paper.
- **The proposed planner:** This planner is the proposed approach MPCFormer. It can be seen as a **PRoactively Socially-aware (PRS)** approach.

*4) Measures of effectiveness*

Measures of effectiveness (MOE) are adopted as follows. **Success / Failure / Collision rate.** Success rate means the percentage of successful off-ramps among 100 random tests in different cases; failure rate means the percentage of missed off-ramps but not colliding; collision rate means the percentage of collisions during the off-ramp; **Vehicle trajectories.** The sampled vehicle trajectories of EAV and SVs used for validating planning socially-aware capability and explainability; **Frobenius norm** of the data-driven social interaction matrices. The higher value of the Frobenius norm explains the higher social interaction strength between vehicles; **Off-ramp duration.** Average length of time required for EAV off-ramp over random tests; **Average speed.** EAV's average speed over random tests; **Actual safety.** The average value of EAV's time headway under a predefined safety threshold is used to measure actual risk; **Perceived safety.** The perceived risk is quantified through frequency-domain analysis [49] of the EAV's time headway. The magnitude of power across frequencies reflects the oscillation intensity and stability of the EAV's behavior; **Speed distribution.** Speed distribution of EAV in different cases; **Lane-change distance.** The longitudinal distance required for one lane-change of EAV in different cases.

*5) Overall planning capability validation*

The results confirm that the proposed planner outperforms the PAS and NS planners, further indicating the advantage of proactive and socially-aware autonomous driving, as evidenced by the off-ramp success rate in TABLE 9. The proposed planner achieves a superior success rate of over 90% across all cases. In comparison, the NS planner outperforms the PAS planner, with an average success rate of 71% versus 39.3%. It demonstrates that the proposed planner has advantages in highly interactive scenarios. Meanwhile, the proposed approach achieves the lowest collision rate with an average of 0.5%. This is attributed to the proposed planner's ability to consider SVs' reactions while considering hard safety constraints. Furthermore, it can be found that all three planners have the highest success rate in case 3 and the lowest success rate in case 4 because case 3 is the most challenging scenario, and case 4 the opposite.

*6) Explainable socially-aware capability validation*

Furthermore, this paper randomly selects a case to analyze the explainable socially-aware planning capability, as shown in Figure 11. In this case, the initial traffic conditions and initial positions of the EAV are identical for all three planners. In Figure 11 (a), the EAV decelerates and returns to the original lane at around 2 s when attempting a lane change. Qualitatively, it can be seen that EAV's Front Vehicle (FV) is also attempting a lane change. EAV is aware of the risk and chooses to avoid it. When the FV drives away, the EAV tries to accelerate and cut in to reach the target lane. This case demonstrates the socially-aware planning capability of the EAV. Furthermore, the social interaction process of Figure 11 (a) is also explainable. TABLE 10 shows the time-variant Frobenius norm value of the data-driven social interaction matrices when EAV interacts with other SVs. It can be seen that when considering social interaction with FV, the value is high and increases in the first 3 seconds, which aligns with the interaction process with FV in Figure 11 (a). The value is also high when interacting with RRV at the 5th second. It makes sense as the EAV is cutting in front of the RRV. TABLE 10 demonstrates that the Frobenius norm value of the data-driven social interaction matrices can explain the social interaction process of the proposed planner.

TABLE 9 The success/failure/collision rate comparison in different cases.

| Case type | Planner type | Success (%) | Failure (%) | Collision (%) |
|---|---|---|---|---|
| Case 1 | PAS planner | 39 | 44 | 17 |
| | NS planner | 72 | 9 | 19 |
| | The proposed planner | **96** | 4 | 0 |
| Case 2 | PAS planner | 31 | 46 | 23 |
| | NS planner | 69 | 9 | 22 |
| | The proposed planner | **94** | 6 | 0 |
| Case 3 | PAS planner | 29 | 49 | 22 |
| | NS planner | 56 | 12 | 32 |
| | The proposed planner | **90** | 8 | 2 |
| Case 4 | PAS planner | 49 | 38 | 13 |
| | NS planner | 82 | 6 | 12 |
| | The proposed planner | **99** | 1 | 0 |
| Case 5 | PAS planner | 49 | 35 | 16 |
| | NS planner | 76 | 8 | 16 |
| | The proposed planner | **95** | 5 | 0 |
| Case 6 | PAS planner | 39 | 46 | 15 |
| | NS planner | 71 | 10 | 19 |
| | The proposed planner | **94** | 5 | 1 |

TABLE 10 The time varied Frobenius norm value of the data-driven social interaction matrices.

| | Frobenius norm value | | | | | |
|---|---|---|---|---|---|---|
| **Social interaction with:** | **FV** | **RV** | **LFV** | **LRV** | **RFV** | **RRV** |
| Time@1s | 1.05 | 0.41 | 0.33 | - | - | 0.40 |
| Time@2s | 1.21 | 0.36 | 0.32 | 0.29 | 0.36 | 0.71 |
| Time@3s | 1.29 | 0.36 | 0.33 | 0.28 | 0.35 | 0.65 |
| Time@4s | 0.65 | 0.37 | 0.99 | 0.29 | 0.36 | 0.76 |
| Time@5s | 0.43 | 0.36 | 0.32 | 0.28 | 0.35 | 0.85 |
| Time@6s | 0.37 | 0.48 | 0.35 | 0.28 | - | - |

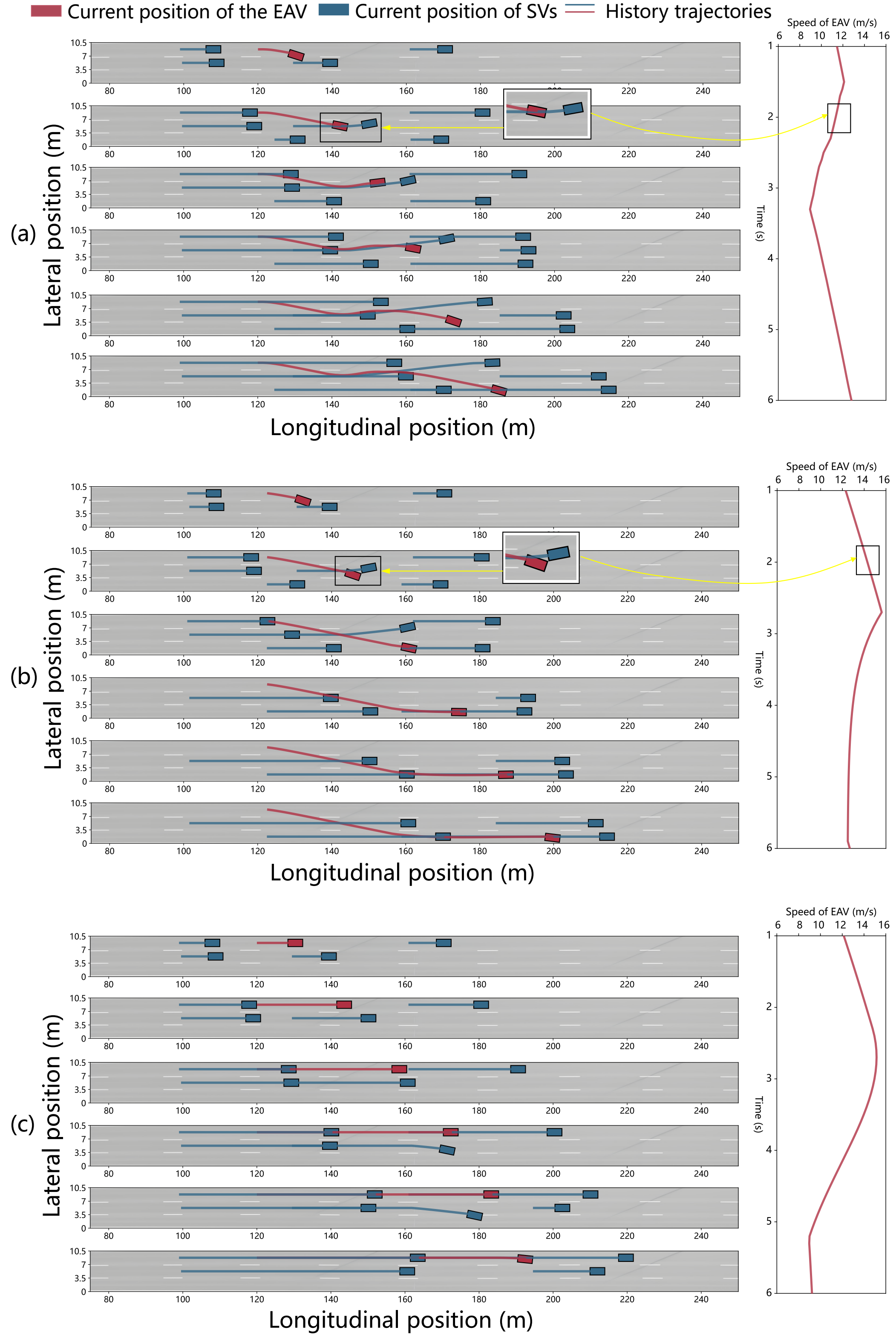

Figure 11 The planning capability comparison in a randomly selected case. (a) the proposed planner; (b) the NS planner; (c) the PAS planner. (The trajectories from top to bottom corresponds to timestep at 1s to 6s.)

In Figure 11 (b), the EAV equipped with the NS planner changes lanes without hesitation. Compared with the proposed planner, at the 2nd second, the NS planner didn't choose to decelerate and nearly resulted in a collision with the FV. This case indicates that the NS planner lacks enough socially-aware planning capability in a complex traffic environment. In Figure 11 (c), the PAS planner was unable to execute a lane change throughout the scenario due to its passive and conservative reaction nature.

*7) Driving efficiency validation*

The results confirm that the proposed planner improves driving efficiency. As illustrated in Figure 12, it reduces off-ramp duration compared with other planners, saving up to 11.23s (9.15s on average) over the NS planner and up to 14.62 s (9.19s on average) over the PAS planner. These improvements stem from proactive interaction-aware motions. In addition, the NS planner achieves shorter durations than the PAS planner in Case 1 to 3, but longer durations in Case 4 to 6, suggesting that the NS planner handles aggressive driving styles better than the PAS planner.

Figure 13 shows the EAV average speed comparison in different cases. The proposed planner improves average speed by 15.75% over the NS planner and 15.23% over the PAS planner. Its speed decreases under higher v/c ratios and aggressive driving due to denser interactions. Although the NS planner shows lower speed than the PAS planner in Case 1 to 3, it achieves shorter off-ramp durations, suggesting more efficient lane changes.

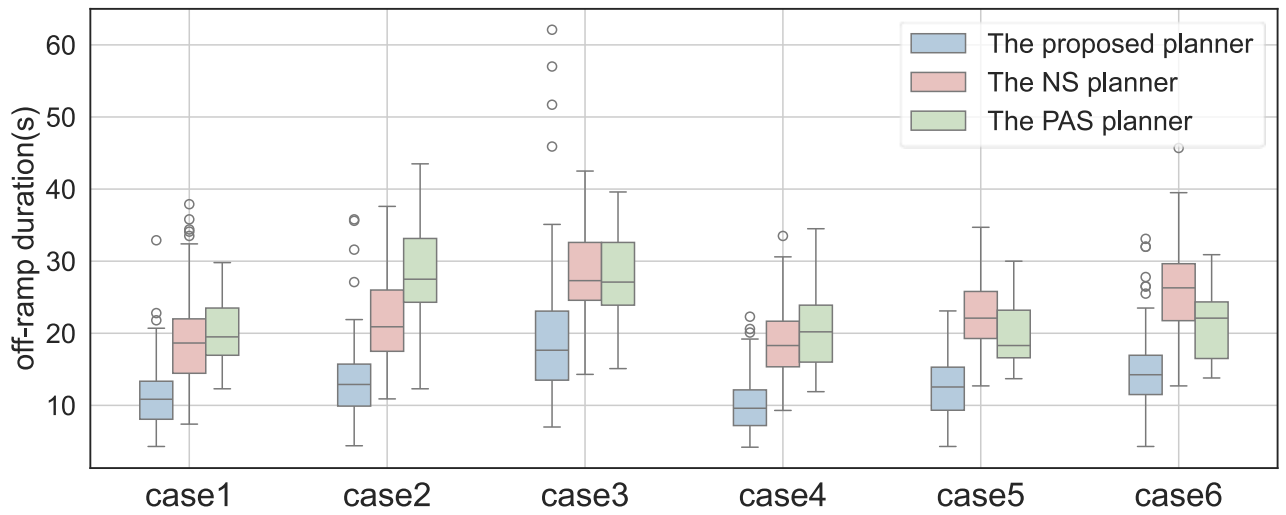

Figure 12 The off-ramp duration of EAV in different cases.

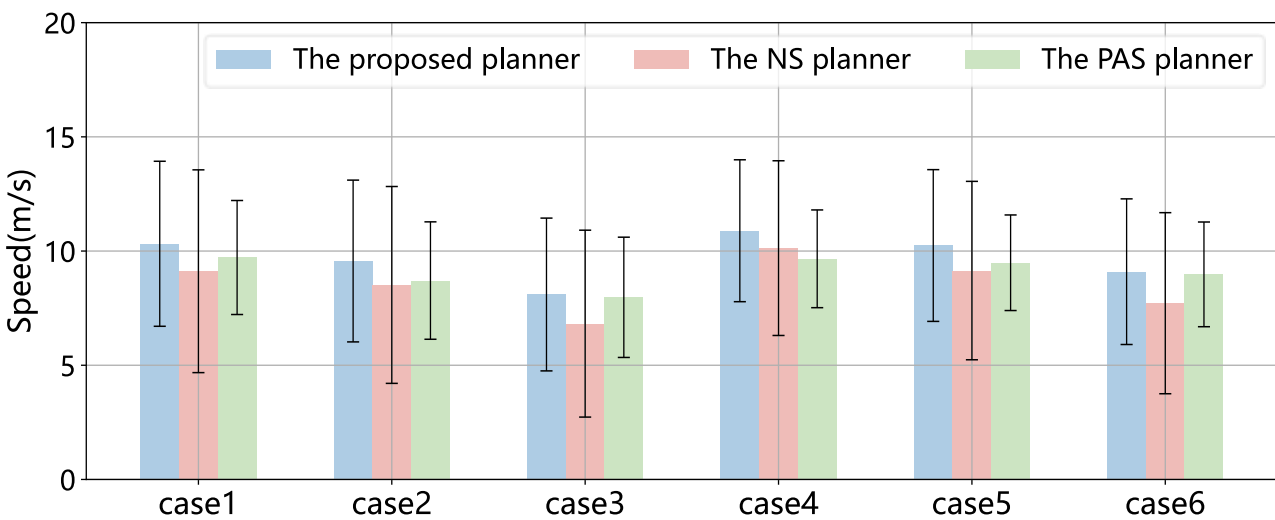

Figure 13 The average speed of EAV in different cases.

*8) Driving safety validation results*

The results confirm that the proposed planner effectively ensures safety. The collision rates shown in TABLE 9 indicate that the proposed planner has the lowest risk to collide. Meanwhile, in those "no-collision" tests, the proposed planner achieves superior actual safety performance by the largest average time headway in Case1(**0.39**, 0.30, 0.34), Case4 (**0.43**, 0.20, 0.43), and Case6 (**0.43**, 0.15, 0.31) as shown in Figure 14. This is because the proposed planner proactively interacts with surrounding vehicles to reduce risky behaviors. However, the PAS planner outperforms in Case2 (0.25, 0.22, **0.31**), Case3 (0.34, 0.20, **0.42**), and Case5 (0.34, 0.27, **0.44**). This can be attributed to its "passive reaction" nature, prioritizing safety with greater confidence, which has benefits in highly interactive scenarios like Case 2 and Case 3. Meanwhile, the NS planner displays the worst safety performance across all cases. This reflects the limitation that the pure learning-based approaches are often hard to guarantee safety.

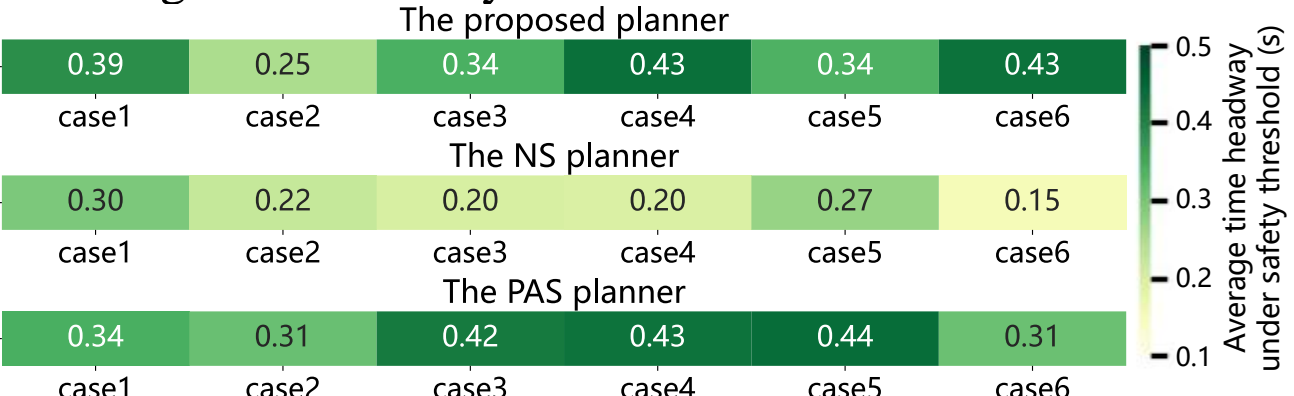

Figure 14 The EAV average time headway comparison.

The perceived safety is evaluated based on the frequency-domain analysis of the EAV's time headway, as illustrated in Figure 15. The analysis is based on the discrete Fourier transform (DFT), which is efficiently computed using the Fast Fourier Transform (FFT) algorithm [49]. Results show that the proposed planner reduces high-frequency oscillations in time headway, indicating smoother and more stable behavior. Compared with the NS and PAS planners, it consistently exhibits lower spectral power above 1 Hz. In contrast, the PAS planner shows significantly higher power in mid-to-high frequency bands, especially in Case 3 and Case 5, suggesting frequent and abrupt adjustments. The NS planner achieves intermediate performance but remains less stable than the proposed planner.

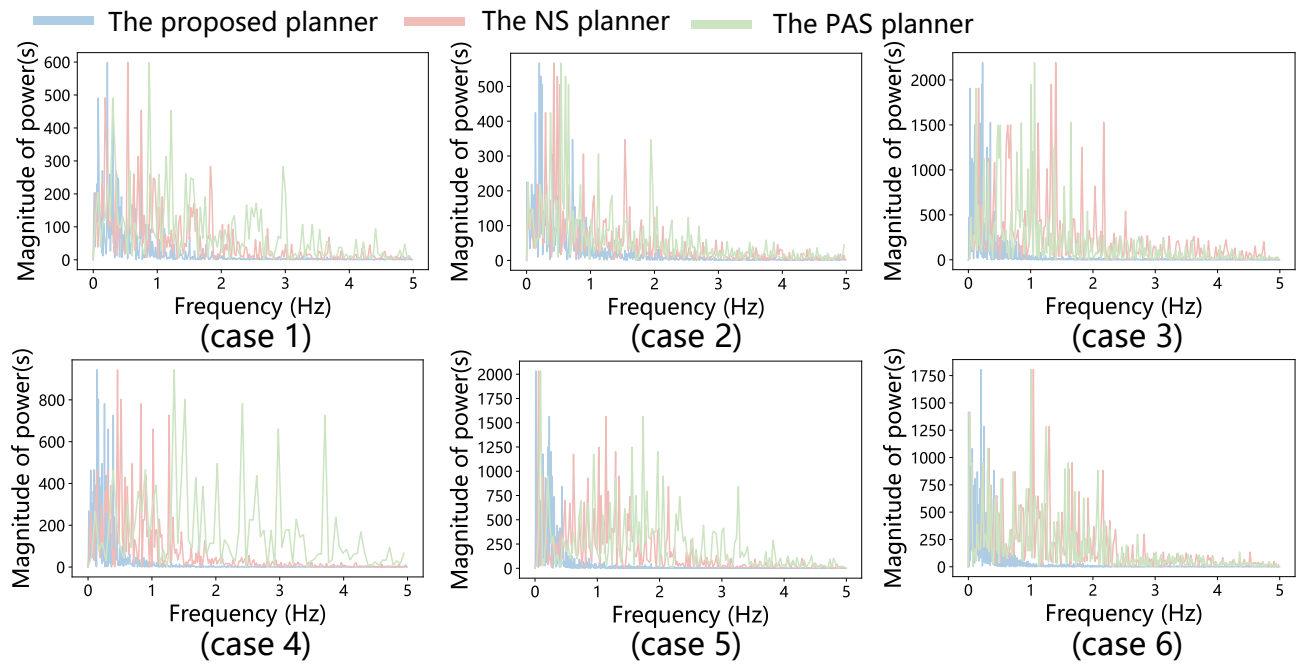

Figure 15 Frequency analysis of EAV's time headway comparison in different cases.

*9) Driving flexibility validation results*

The results confirm that the proposed planner demonstrates high flexibility in the highly interactive traffic scenarios. Figure 16 compares the lane change distance distribution across different cases. The commonality among the six cases is that the PAS planner obviously has a larger lane change distance. The proposed planner saves 49.01% of lane change distance on average compared with the PAS planner. As the PAS planner relies on predefined rules and passively reacts to surrounding vehicles, resulting in struggling in interactive scenarios. Compared with the NS planner, the proposed planner has a wider box of the lane change distribution in Case 1 to Case 3, but the opposite in Case 4 to Case 6. This indicates the proposed planner is more competitive and can adjust its lane change distance flexibly to obtain better planning capability in aggressive traffic.

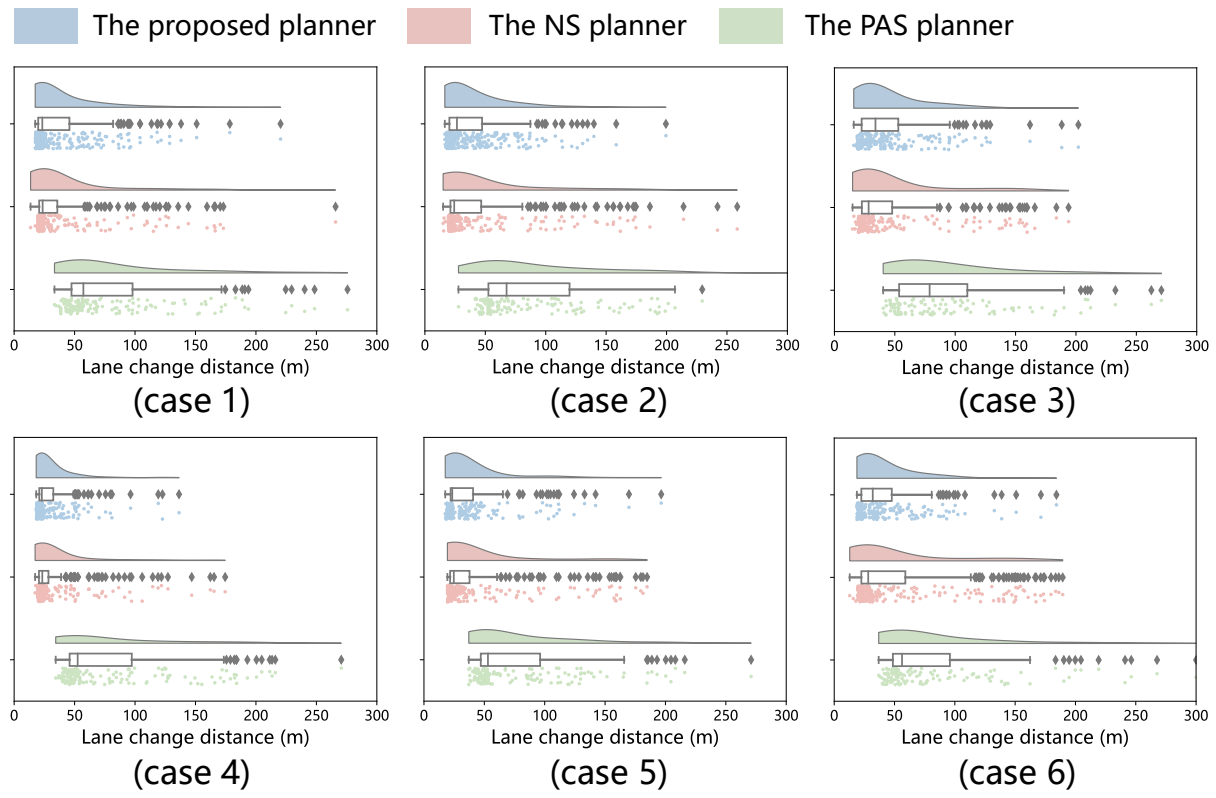

Figure 16 The lane change distance comparison in different cases.

### 10) *Inference latency analysis*

The inference latency results in TABLE 11 demonstrate that our approach satisfies real-time requirements, with total inference latency consistently below the planning interval (0.2 seconds). The Transformer-based encoder–decoder maintains stable latency (50–60 milliseconds), as its main computational cost stems from processing historical trajectories and local map features, which are independent of the future planning horizon. In contrast, the MPC planner's latency scales with the planning horizon, ranging from 24.9 to 95.0 milliseconds.

TABLE 11 The inference latency analysis under different future planning horizons. (NVIDIA Geforce RTX 3050 Laptop GPU, and 12th Gen Intel(R) Core(TM) i5-12500H CPUs)

| **Future planning horizon (s)** | **Planning interval (ms)** | **Inference latency (ms)** | | |
|---|---|---|---|---|
| | | **Encoder-decoder** | **MPC planner** | **Total** |
| 5 | 200 | 59.6 | 95.0 | 154.6 |
| 3 | | 57.4 | 61.0 | 137.4 |
| 1 | | 51.1 | 24.9 | 76.0 |

## IV. Conclusion

This paper introduces ***MPCFormer***, an explainable socially-aware autonomous driving approach with physics-informed and data-driven coupled social interaction dynamics. In this model, the dynamics are formulated into a discrete space-state representation, which embeds physics priors to enhance modeling explainability. The dynamics coefficients are learned from naturalistic driving data via a Transformer-based encoder-decoder architecture. To the best of our knowledge, MPCFormer is the first approach to explicitly model the dynamics of multi-vehicle social interactions. By integrating the learned social interaction dynamics into MPC planning based on a leader-follower game framework, the AD vehicle can generate manifold, human-like behaviors when interacting with surrounding traffic while mitigating the potential safety risks typically associated with purely learning-based approaches. Open-looped prediction and close-looped planning results demonstrate that:

- In open-loop prediction, the proposed MPCFormer achieves the lowest SV trajectory prediction error among state-of-the-art approaches, achieving an ADE as low as 0.86 m over a long prediction horizon of 5 seconds.
- In close-looped planning, the proposed MPCFormer ensures the highest overall planning capability with an average success rate of 94.67%. The proposed MPCFormer reduces the collision rate from 21.25% to 0.5%. It also has a validated explainable socially-aware planning capability.
- The proposed MPCFormer also demonstrates efficiency gains of up to 15.75% and 15.23% compared to the two baseline planners and ensures perceived and actual safety. Furthermore, the proposed MPCFormer is more competitive in interactive scenarios with a smaller lane-change distance.

The limitations of this work lie in the lack of consideration of the V2X communications and the explicit driving styles of SVs. Besides, it should be noted that the current evaluation is conducted on the NGSIM highway dataset. Although highway scenarios involve relatively structured interactions, the proposed framework is not restricted to highway-specific assumptions. Since the social interaction matrices are fully data-driven and conditioned on vehicle trajectories and local map representations, the framework is, in principle, extensible to more complex urban environments. Extending validation to large-scale urban datasets will be part of future work.

## Appendix

### *Quantitative analysis of the linearization error of the vehicle kinematic model*

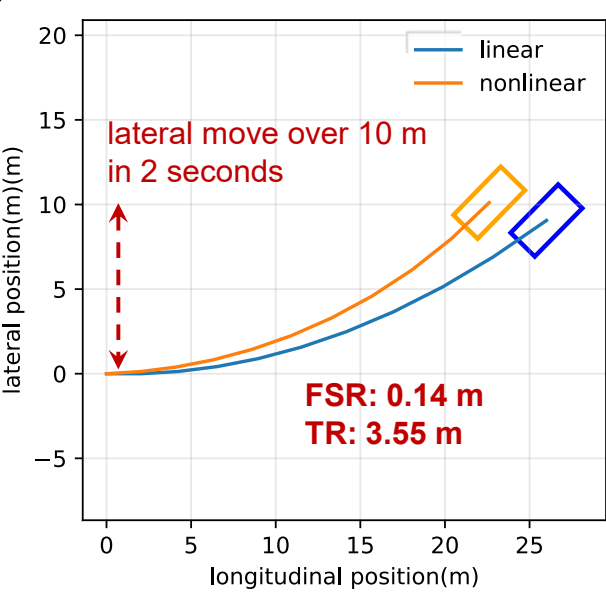

Figure 17 The trajectory analysis of the nonlinear and linear model (FSR: first step L1 error, TR: terminal L1 error).

To assess the small-angle linearization error, we compare the nonlinear (Eq. 5) and linearized kinematic bicycle models under a high-curvature maneuver. The horizon is 2 s with a 0.2 s interval, maximum acceleration and steering inputs (3 m/s² and 0.15 rad), and an initial speed of 10 m/s. As shown in Figure 17, the vehicle achieves over 10 m lateral displacement within 2 s, while the first-step L1 error (FSR) is only 0.14 m. Since MPC re-plans every 0.2 s, errors do not accumulate across steps, indicating that the linearization error remains acceptable for motion planning.

### *Discussion on the robustness of the proposed approach*

The proposed approach is expected to remain robust to noise for two reasons. First, receding-horizon MPC continuously updates vehicle states and replans, limiting error accumulation. Second, the Transformer encoder–decoder is trained on real-world trajectories with measurement noise, enhancing robustness. Future work will further explore explicit noise modeling and stochastic MPC formulations.

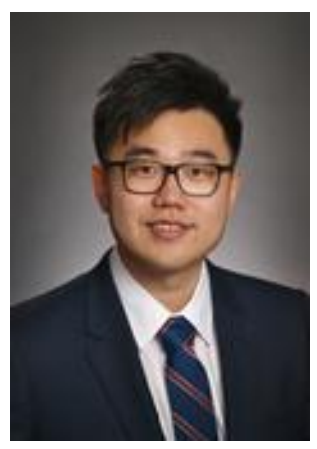

**Jia Hu** (Senior Member, IEEE) is currently working as a Zhongte Distinguished Chair of Cooperative Automation with the College of Transportation Engineering, Tongji University. Before joining Tongji University, he was a Research Associate with the Federal Highway Administration (FHWA), USA. He is an Editorial Board Member of the *Journal of Intelligent Transportation Systems* and the *International Journal of Transportation Science and Technology*. He is a member of TRB (a Division of the National Academies) Vehicle Highway Automation Committee, the Freeway Operations Committee, Simulation subcommittee of Traffic Signal Systems Committee, and the Advanced Technologies Committee of the ASCE Transportation and Development Institute. He is the Chair of the Vehicle Automation and Connectivity Committee of the World Transport Convention. He is an Associate Editor of the American Society of Civil Engineers *Journal of Transportation Engineering* and IEEE OPEN JOURNAL OF INTELLIGENT TRANSPORTATION SYSTEMS.

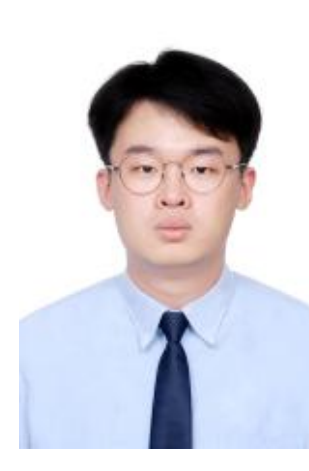

**Zhexi Lian** was born in Shanxi, China. He received the bachelor's degree in transportation engineering from College of Transportation, Tongji University, Shanghai, China, in 2023. He is currently pursuing the Ph.D. degree with Key Laboratory of Road and Traffic Engineering of the Ministry of Education, Tongji University. His main research interests include data-driven control, end-to-end autonomous driving and vision language action models.

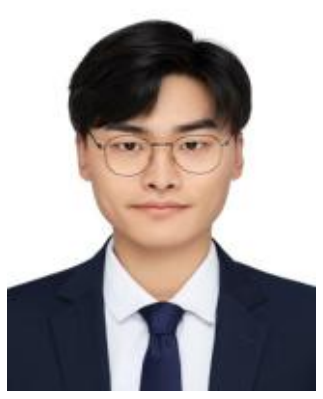

**Xuerun Yan** received his B.S. degree in traffic engineering from Southeast University in 2021. He currently works as a research assistant with the Key Laboratory of Road and Traffic Engineering, Ministry of Education, Tongji University, Shanghai, China. His research interests are in the areas of AV simulation, AV planning, and cooperative driving.

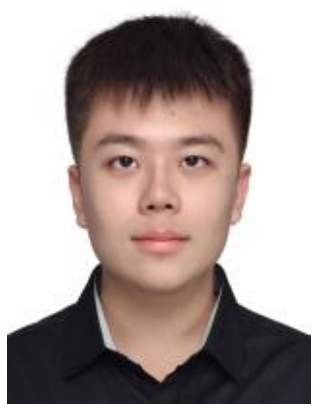

Ruiang Bi received the bachelor's degree in transportation engineering from Tongji University, Shanghai China, in 2024. He is currently working toward the master's degree with Tongji University, Shanghai, China. His main research interests include cooperative automation, optimal control, decision making, and behavioral planning.

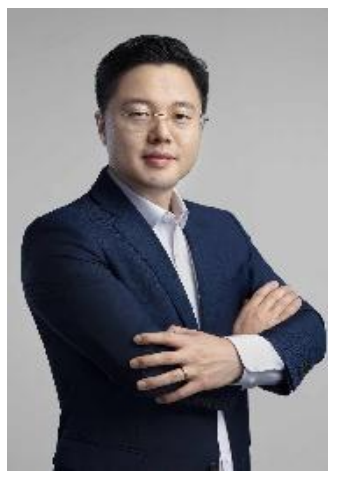

**Dou Shen** received a Ph.D. from the Hong Kong University of Science and Technology, and currently serves as executive vice president of Baidu and the president of Baidu AI Cloud Group. Dr. Shen joined Baidu in 2012 and has served in various management roles, including web search, display advertising, the financial services group and mobile products. Dr. Shen has published more than 40 papers in international conferences and journals, and held multiple patents on Internet search and computational advertising. Currently, he serves as the Vice President of SIGKDD China Chapter, the Dean of the School of Artificial Intelligence at North China Electric Power University, the independent director of COSCO Shipping Holdings Co., Ltd., and the director of companies such as China Unicom, CITIC AIBANK, Neusoft, and iQiyi.

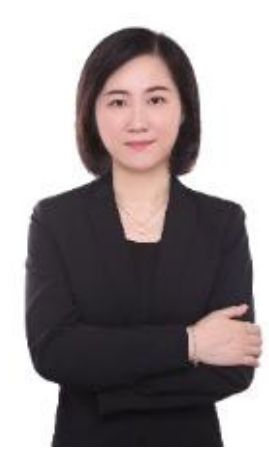

**Yu Ruan** graduated from Ningxia University and holds an EMBA from China Europe International Business School (CEIBS). She is currently Vice President of Baidu and General Manager of Baidu Intelligent Cloud Application Product Division. With over a decade of experience in Internet product operations, Yuan Yu is an industry expert in product management, operations, and brand marketing. She has extensive management experience and has led her teams to achieve multiple business breakthroughs and innovative transformations.

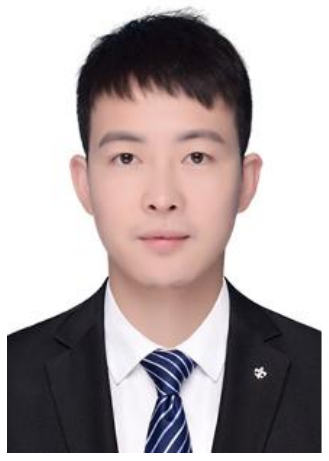

**Chunlong Xia** received the B.S. degree in automation from Southwest Jiaotong University, China, in 2016, and the M.S. degree from the Institute of Artificial Intelligence and Robotics at Xi'an Jiaotong University in 2019. He is currently the head of algorithms in the field of intelligent transportation vision at Baidu Inc., where he focuses on research, implementation, and the development of industry standards for large model algorithms in traffic perception.

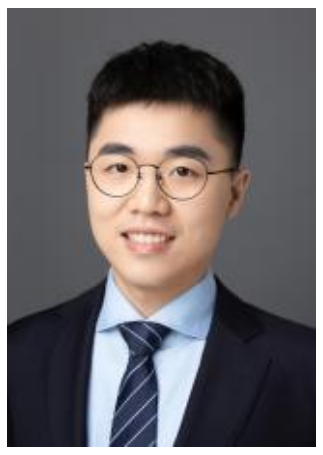

**Haoran Wang** received the bachelor's degree in transportation engineering from Tongji University, Shanghai, China, in 2017, and the Ph.D. degree from Tongji University in 2022. He is currently a Postdoctoral Researcher with the College of Transportation Engineering, Tongji University. He is a researcher on vehicle engineering, majoring in intelligent vehicle control and cooperative automation. Dr. Wang served the IEEE TRANSACTIONS ON INTELLIGENT VEHICLES, IEEE TRANSACTIONS ON INTELLIGENT TRANSPORTATION SYSTEMS, *Journal of Intelligent Transportation Systems*, and IET *Intelligent Transport Systems* as peer reviewer with a good reputation.